\documentclass{article}

\PassOptionsToPackage{numbers, compress}{natbib}
 \usepackage[final]{nips_2016}

\usepackage[utf8]{inputenc} 
\usepackage[T1]{fontenc}    
\usepackage{hyperref}       
\usepackage{url}            
\usepackage{booktabs}       
\usepackage{amsfonts}       
\usepackage{nicefrac}       
\usepackage{microtype}      

\usepackage{graphicx} 
\usepackage{subfigure} 
\usepackage{algorithm}
\usepackage{algcompatible}

\usepackage{amsmath}
\usepackage{amssymb} 
\usepackage{amsfonts}
\usepackage{mathtools}
\usepackage[small,compact]{titlesec}

\usepackage{color}

\definecolor{gray}{rgb}{0.7,0.7,0.7}

\newcommand{\expectationE}[2]{ \mathbb{E}_{#2}  \left[ #1 \right] }

\DeclarePairedDelimiter\floor{\lfloor}{\rfloor}

\title{Learning and Transfer of Modulated Locomotor Controllers}

\author{Nicolas Heess, \ Greg Wayne, \ Yuval Tassa, \ Timothy Lillicrap, \ Martin Riedmiller, \ David Silver \\
Google DeepMind
}

\begin{document}

\maketitle

\begin{abstract}
We study a novel architecture and training procedure for locomotion tasks. A high-frequency, low-level ``spinal'' network with access to proprioceptive sensors learns sensorimotor primitives by training on simple tasks. This pre-trained module is fixed and connected to a low-frequency, high-level ``cortical'' network, with access to all sensors, which drives behavior by modulating the inputs to the spinal network. Where a monolithic end-to-end architecture fails completely, learning with a pre-trained spinal module succeeds at multiple high-level tasks, and enables the effective exploration required to learn from sparse rewards. We test our proposed architecture on three simulated bodies: a 16-dimensional swimming snake, a 20-dimensional quadruped, and a 54-dimensional humanoid (see attached \href{https://youtu.be/sboPYvhpraQ}{video}).
\end{abstract}

\section{Introduction}

A newborn antelope attempts its first steps only minutes after birth and is capable of walking within half an hour. A human baby, when lifted so that its feet just graze the ground, will step in a cyclical walking motion \citep{peiper1929schreitbewegungen, dominici2011locomotor}. Nature provides clear examples of evolutionarily determined, innate behaviors of surprising complexity, and basic locomotor circuits are well-developed prior to any significant goal-directed experience. The innate structure simplifies the production of goal-directed behavior by confining exploration to stable and coherent, yet flexible dynamics.

On the other hand, machine learning's recent success stories have emphasized rich models with weakly-defined prior structure, sculpted by large amounts of data. And indeed, several recent studies have shown that end-to-end reinforcement learning approaches are capable of generating high-quality motor control policies using generic neural networks \citep{schulman2015high, levine2015end, heess2015learning, lillicrap2015continuous, gu2016continuous}.  A key question is how to get the best of both worlds: to build modular, hierarchical components that support coherent exploration while retaining the richness and flexibility of data-driven learning.

Here, we aim to create flexible motor primitives using a reinforcement learning method. We explore an approach based on a general-purpose, closed-loop motor controller that is designed to be modulated by another system. When modulated by random noise, this low-level controller generates stable and coherent exploratory behavior. A higher-level controller can then recruit these motor behaviors to simplify the solution of complex tasks. As a result, it can learn effective control strategies given only weak feedback, including tasks where reward is only provided infrequently after completion of a goal (a natural description of many tasks).

Our architecture takes inspiration from the division of labor in neurobiological motor control~\citep{kandel2000principles}. Biological motor primitives are formed by spinal cord interneurons, which mediate both fast, reflexive actions and the elementary constituents of motor behaviors like locomotion and reaching. The spinal cord has direct sensory inputs that measure muscle tension and load, joint position, and haptic information from the skin. These sensory modalities are \emph{proprioceptive} (``taken from near'') because they carry information measured at close range to the body as opposed to the \emph{exteroceptive} (``taken from afar'') modalities like vision, audition and olfaction. Cortical motor neurons produce voluntary motor behavior primarily through the modulation of these interneuron populations. 

\vspace{5mm}
Based on these considerations, we explore hierarchical motor controllers with the following properties:

\begin{enumerate}
\item \emph{Hierarchy}: The controller is subdivided into a low-level controller, which computes direct motor commands (e.g. joint torques), and a high-level controller, which selects among abstract motor behaviors.  
\item \emph{Modulation}: The high-level controller outputs a signal that modulates the behavior of the low-level controller, through a communication bottleneck.
\item \emph{Information hiding}: The low-level controller has direct access only to task-independent, proprioceptive information. For the control problems considered in this paper, this proprioceptive information contains, for instance, the joint angles and velocities of the body and haptic information. Notably, it does not include information about absolute position or orientation in space or task-specific information such as a goal location.
The high-level controller has access to all necessary proprioceptive and exteroceptive information.
\item \emph{Multiple time scales}: While the low-level controller operates at a basic control rate (i.e., it receives an observation and produces an action at every time step in the simulation), the high-level controller can operate at a slower rate, updating the modulatory control signal to the low-level controller less frequently.
\end{enumerate}  

These design decisions are intended to create an abstraction barrier between the low and high levels: Separating the high level from the physics and detailed demands of motor actuation and likewise, sheltering the low-level controller from specific task objectives so it can acquire domain-general functionality. 
Below we present results demonstrating that our architecture solves several non-trivial control problems. A more detailed analysis of the implications of the design decisions listed above is, however, left for future work.


\label{sec:Methods:Overview}

\begin{figure}
  \begin{minipage}[c]{0.47\textwidth}
    \includegraphics[width=\textwidth]{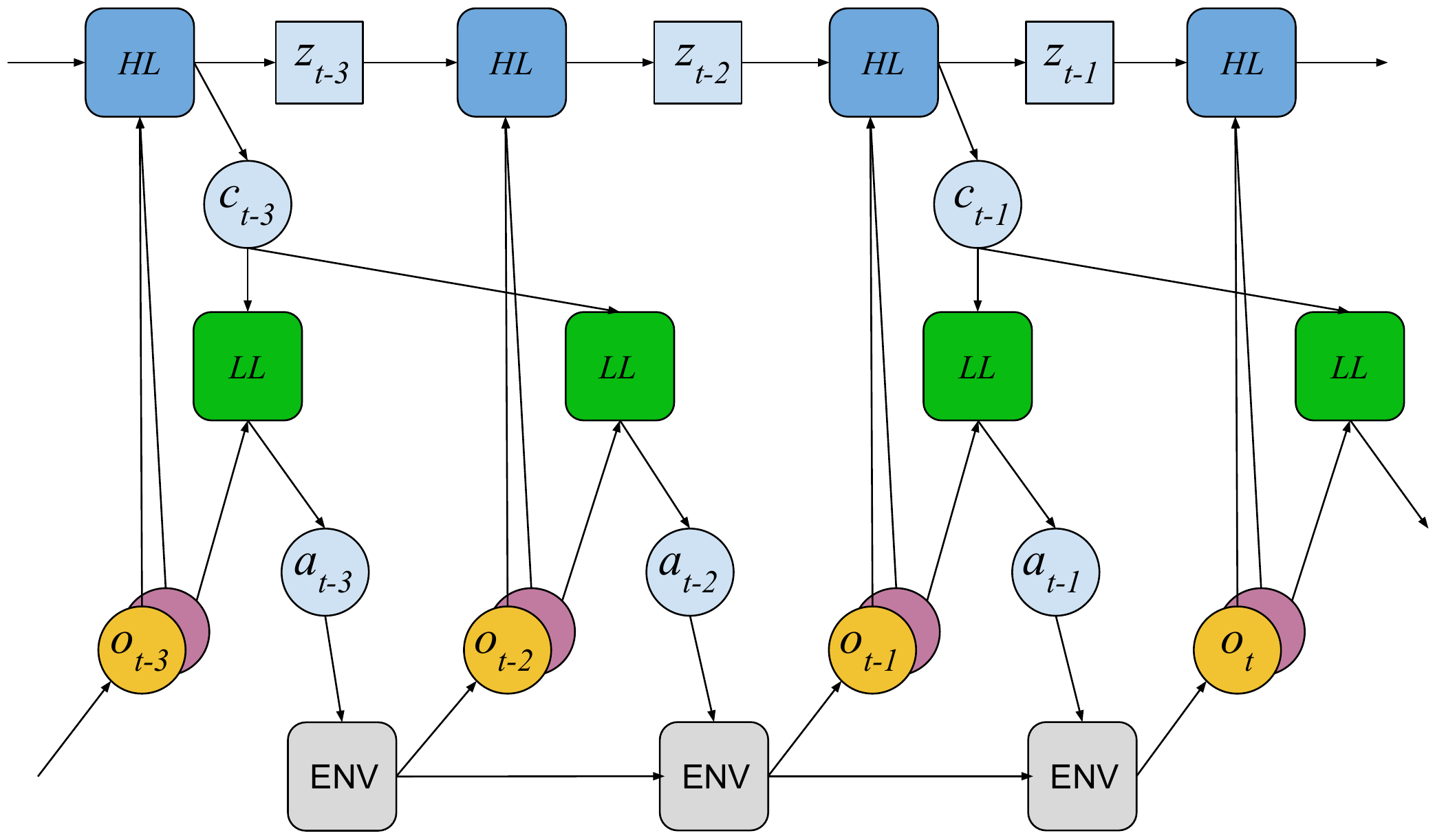}
  \end{minipage}\hfill
  \begin{minipage}[c]{0.5\textwidth}
    \caption{\footnotesize
Structure of our control hierarchy, consisting of the recurrent high-level controller (\emph{HL}, dark blue) and  feedforward low-level controller (\emph{LL}, green). Rounded squares represent functions; squares represent deterministic variables; circles represent stochastic variables. The low-level controller only has access to the proprioceptive component of the current observation ($o_t$, red). The high-level controller has access to all observations ($o_t$, yellow and red). While both controllers observe sensory input at the world-clock frequency, the modulatory control signal $c_t$ from the high-level is updated every $K$ steps (here $K=2$).
    } \label{fig:model}
  \end{minipage}
\end{figure}

\section{Architecture}

We now describe the architecture of the hierarchical controller in greater detail (see Fig.\ 1). The setup is the standard agent-environment interaction model. At each point in time $t$, an agent executes an action $a_t$, and subsequently receives a reward $r_t$ and a new observation $o_{t+1}$. Its goal is to maximize the expected sum of discounted future rewards $R_t = \sum_{t'=t  } ^\infty \gamma^{t' - t} r_{t'}$, known as the \emph{return}. 

The agent is characterized by a policy $\pi$ with parameters $\theta$ which specify a distribution over actions as a function of the history $h_t = ( o_1, a_1, \dots o_{t-1}, a_{t-1}, o_t )$, i.e.\ $a_t \sim \pi(\cdot | h_t; \theta).$\footnote{In this work, we did not find it necessary to include the actions in the history and left them out.} The stochastic policy is defined by the composition of networks for the high-level controller $F_H$ and low-level controller $F_L$. This combined network outputs the parameters of the action distribution $\pi$. In the experiments below, actions are multi-dimensional, and we parameterize the action distribution as a factorized Normal distribution with mean and variance being functions of $h_t$: $a_t \sim \pi(\cdot | h_t) = \mathcal{N}(\cdot | \mu(h_t), \sigma^2(h_t))$.

In the experiments presented here, the low-level controller is a non-recurrent neural network that maps the proprioceptive information $o^P$ and the control signal received from the high-level controller $c$ onto the parameters of the action-distribution:
\begin{align}
( \mu, \sigma) &= F_L( o^P, c) \\
a & \sim \mathcal{N}( \cdot | \mu, \sigma^2).
\end{align}
For the high-level controller, we have used recurrent networks $F_H = ( f_H, g_H)$ that integrate observations at every time step and produce a new control signal $c_t$ every $K$ time steps:
\begin{align}
z_t &= f_H(o_t^F, z_{t-1}) \\
c_t &= g_H( z_{{\tau} (t)})  \\
\tau(t) &= \floor{(t - 1)/ K} K+1
\end{align}
where $o_t^F$ is the full observation (including task-specific information), $z_t$ is the recurrent state of the high-level controller, $K$ is the control interval, and $\tau(t)$ is the most recent update time for the high-level control signal.

\section{Learning locomotor controllers with policy gradients}
\label{sec:Methods:PG}

We use an actor-critic policy gradient framework for learning during pre-training as well as transfer. We consider both fully observed (MDPs) and partially observed problems (POMDPs). 
\subsection{Generalized advantage estimation} We perform gradient ascent in the expected discounted return $J = \expectationE{R_0}{}$, where the expectation is taken with respect to the trajectory distribution induced by the policy and the environment dynamics.

This gradient is given by
\begin{equation}
\nabla_\theta J = \sum_t \expectationE{ \nabla_\theta \log \pi(a_t | h_t; \theta) ( R_t - b_t)}{}, \label{eq:ReinforceUpdate}
\end{equation}
where $b_t$ is some baseline that does not depend on $a_{t' \geq t}$. In this work we use a learned, parameterized value function $V_t(h_t; \omega)$ with parameters $\omega$ to lower the variance of the estimate. We replace $R_t$ by the $\lambda$-weighted return $R^\lambda_t = \sum_{k=0}^\infty \lambda^k R_t^k$
where $R_t^k = \sum_{j=0}^k \gamma^j r_{t+j} + \gamma^{t + k + 1} V(h_{t+k+1})$. The parameter $\lambda$ trades off bias in the value function against variance in the return. We also use estimates of the return $R^{\lambda'}_t$ as targets for the value function, so that the loss for value function training is
\begin{equation}
L(\omega) = \frac{1}{2} \sum_t ||R^{\lambda'}_t - V(h_t; \omega)||^2.
\end{equation}
Note that we allow for different values of $\lambda$ and $\lambda'$ for computing the policy and value function updates, respectively. Additionally, although each $R^{\lambda'}_t$ nominally includes value function terms dependent on $\omega$ from future time steps, we do not differentiate with respect to them, as is typical for temporal difference learning.

\subsection{Policy gradient with hierarchical noise}
The policy gradient framework outlined above performs on-policy learning where the stochasticity of the policy is used for exploration. Choosing the action distribution $\pi$ to be a diagonal Gaussian is common due to its simplicity. At the same time, as we will demonstrate below, it can lead to very poor exploratory behavior, especially in high-dimensional action spaces. Due to its restricted form, it is unable to describe correlations across action dimensions or time steps. Actuating physical bodies with this form of white noise tends to produce undirected, twitchy movements that are attenuated by the second-order dynamics of the physics.

In contrast, our low-level controllers are feedback controllers that produce pre-trained locomotor behavior. Modulating this behavior appropriately can lead to exploratory behavior that is more consistent in space and time.
Thus, we allow stochasticity not only at the output of the low-level controller but also in the high-level controller. More precisely, in transfer training we treat the high-level controller as a stochastic network where 
\begin{align}
c_t  &=  \tilde{c}_{\tau(t)} \\
\tilde{c}_{\tau(t)} &\sim \pi^H(\cdot | z_{\tau(t)}) = \mathcal{N}(\mu^H( z_{\tau(t)} ), \sigma^H( z_{\tau(t)} ) ).  \label{eq:stochastHLExplicit}
\end{align}
The policy distribution composed of the high- and low-level controllers can be seen as a distribution with latent variables: $\pi(a_t | h_t ) = \int \pi( a_t | h_t, \tilde{c}_{\tau(t)}) \pi^H(  \tilde{c}_{\tau(t)} | h_{\tau(t)}, z_{\tau(t)}) \mathrm{d} \tilde{c}_{\tau(t)}$. 

This hierarchical model can be optimized in different ways. The particular approach we take relies on the re-parameterization trick recently applied in the probabilistic modeling literature \cite{kingma2013,rezende2014} and in a policy gradient framework by \cite{heess2015learning}:

To use the re-parameterization trick, note that an equivalent formulation of equation (\ref{eq:stochastHLExplicit}) can be obtained by considering $c_t = \tilde{g}^H( z_{\tau}, \epsilon_{\tau})$, where $\tilde{g}^H( z_{\tau}, \epsilon_{\tau}) = \mu^H(z_{\tau}) + \sigma^H(z_{\tau}) \epsilon_{\tau}$ and $\epsilon_{\tau} \sim \mathcal{N}(0, \mathbf{I})$. In this view, $\tilde{g}^H$ is a deterministic function that takes as additional input a random variable drawn from a fixed distribution. Since we have knowledge of $\epsilon_{\tau}$, we can now evaluate $\nabla_\theta \log \pi(a_t | z_\tau, \epsilon_{\tau})$, which would otherwise be difficult for a latent variable model. The policy gradient estimate in equation ($\ref{eq:ReinforceUpdate}$) is simply formed by backpropagating directly into the high-level controller.\footnote{
Using a value function for bootstrapping in combination with hierarchical noise requires extra care since $c_t$ affects future primitive actions. This could be accounted for by making $V$ dependent on $c_t$, or by bootstrapping only after resampling $c_t$. In our experiments we ignore this influence on the value and use $V(h_t; \omega)$ as above.
}

This high-level noise can achieve a dramatically different effect from i.i.d.\ noise added to each action dimension independently at every time step. Since the high-level noise is held constant over the high-level control interval and transformed by the low-level controller, it induces spatially and temporally correlated stochasticity at the primitive action level. 


\section{Experiments}

We evaluate our framework on three physical domains: a swimming snake, a quadruped and a humanoid. The snake has 6-links with a 5 dimensional action space and can propel itself forward by exploiting frictional forces. The quadruped has a 8 dimensional action space with two joints per leg. The humanoid has 21 action dimensions. For the following motor control problems, the core challenge is to learn basic locomotion. For more complex behaviors like navigation, the locomotion pattern can be reused. In addition to the description below, we encourage the reader to watch the supplemental videos\footnote{High-quality version at \url{https://youtu.be/sboPYvhpraQ}}.

\subsection{Training methodology}

We train our hierarchical motor controller on a simple \emph{pre-training} task, and then evaluate its performance in one or more \emph{transfer} tasks. This involves training the low-level controller (which will be re-used later) jointly with a provisional high-level controller which provides task-specifc information during pre-training and hence ensures controllability of the low-level controller. The pre-training task, which facilitates the development of generic locomotion skills, is described by an informative shaping reward and requires the controller to move each creature from a random initial configuration to a randomly positioned target. 

After pre-training the provisional high-level controller is discarded and the weights of low-level controller are frozen. For each transfer task a new high-level controller is trained to modulate the inputs of the frozen low-level controller.
The transfer tasks are most naturally described by \emph{sparse} reward functions which are zero everywhere except at goal states. In order to obtain any reward, these tasks demand temporally-extended, structured exploration, posing a significant challenge for reinforcement learning methods. We use multiple transfer tasks for each domain to test the versatility of the learned low-level controllers.

We implemented our experiments using the asynchronous actor-critic framework introduced in~\cite{mnih2016async}. For each experiment and architecture (pre-training and transfer), we perform a coarse grid search over the following hyper-parameters: learning rate, relative scaling of learning rate for the value function, $\lambda$, $\lambda'$, and the length of the backpropagation-through-time truncation window. Unless noted otherwise we report results for the hyper-parameter setting which performed best over an average of 5 repeated experiments.

Depending on the transfer task we compare learning with the pre-trained low-level controller to learning a feedforward (\emph{FF}) or recurrent policy (\emph{LSTM}) from scratch; and we also compare to re-using a pre-learned FF network where we only learn a new input layer (\emph{init FF}; the new input layer is to account for the fact that the observation space may change between pre-training and transfer task, or that the new task requires a different mapping from observations to motor behavior).

\subsection{6-link snake}

\textbf{Pre-training task} The pre-training task initializes the snake at an origin with a random orientation and a random configuration of the joint angles. It is required to swim towards a fixed target over 300 time-steps, which requires being able to turn and swim straight. The low-level controller's sensory input consists of the joint angles, the angular velocities, and the velocities of the 6 segments in their local coordinate frames. The provisional high-level controller is also exposed to an egocentric (i.e., relative to the body frame) representation of the target position. A shaping reward in the form of the negative of the distance to the target is given at every step. The modulatory input from the high-level controller to the low-level controller is updated every $K=10$ time steps.

\begin{figure}
\begin{center}
\includegraphics[width=1\textwidth]{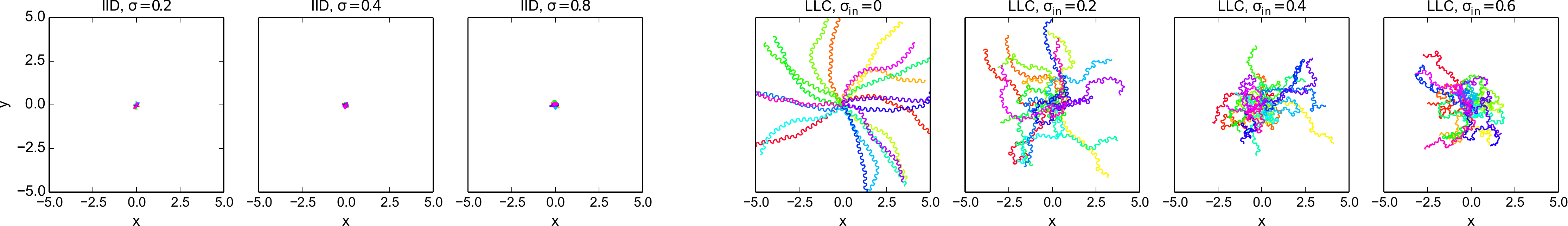}
\end{center}
\caption{\footnotesize
Exploration behavior of the snake with (a) zero mean i.i.d.\ Gaussian noise with three different standard deviations (left three plots) or (b) with the pre-trained low-level controller (LLC) periodically modulated by i.i.d.\ Gaussian noise (right four plots). Each plots shows 4000-step trajectories of the $(x,y)$-position of the anterior segment of the snake after initialization at the origin in a random configuration. Without high-level modulatory input $\sigma_{\mathrm{in}}  = 0$ the snake swims mostly straight; high-level modulation leads to more diverse trajectories.
}
\label{fig:SwimmerNoise}
\end{figure}

\textbf{Analysis of the locomotor primitives} The training task is easily solved. To assess the locomotor primitives embodied by the learned low-level controller, we remove the high-level controller and modulate the low-level controller with i.i.d.\ Gaussian noise sampled every 10 steps. The resulting behavior, obtained by initializing the swimmer at the origin in a random configuration and running for 4000 time steps, is shown in Figure \ref{fig:SwimmerNoise}. Using the center of the most anterior body segment as a reference point, swimming trajectories are shown. Clearly, the locomotor primitives produce coherent swimming behavior, and the nature of the input noise determines the structure of the behavior. In the absence of high-level modulation, the snake swims nearly straight; increasing the amplitude of modulatory noise leads to more variable trajectories. For comparison, we also show the behavior elicited by the commonly used zero-mean i.i.d.\ Gaussian noise applied directly as motor commands, which produces barely any displacement of the body, even at high noise levels. The largest standard deviation we tested was 0.8, which is large compared to the action range $[-1, 1]$.

\begin{figure}
\begin{center}
\includegraphics[width=.27\textwidth]{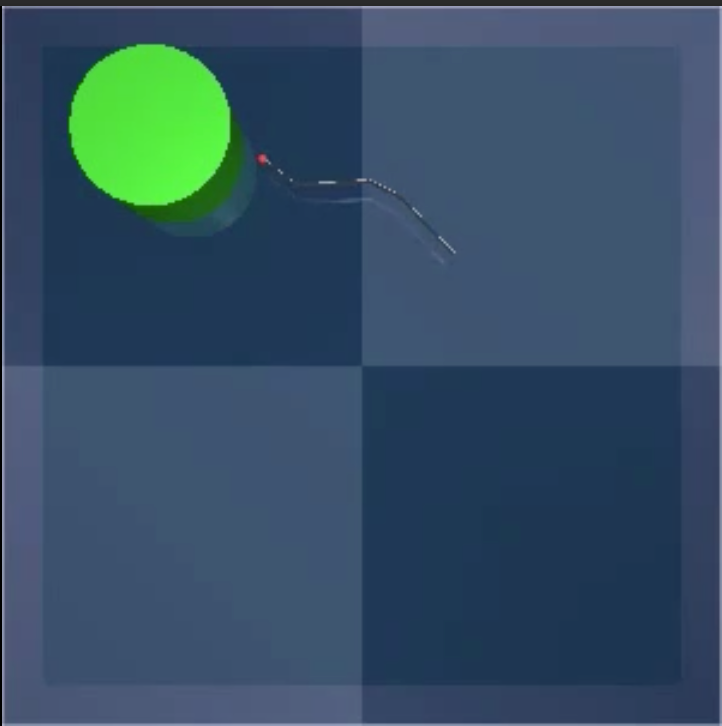}
\hspace{0.1\textwidth}
\includegraphics[width=.3\textwidth]{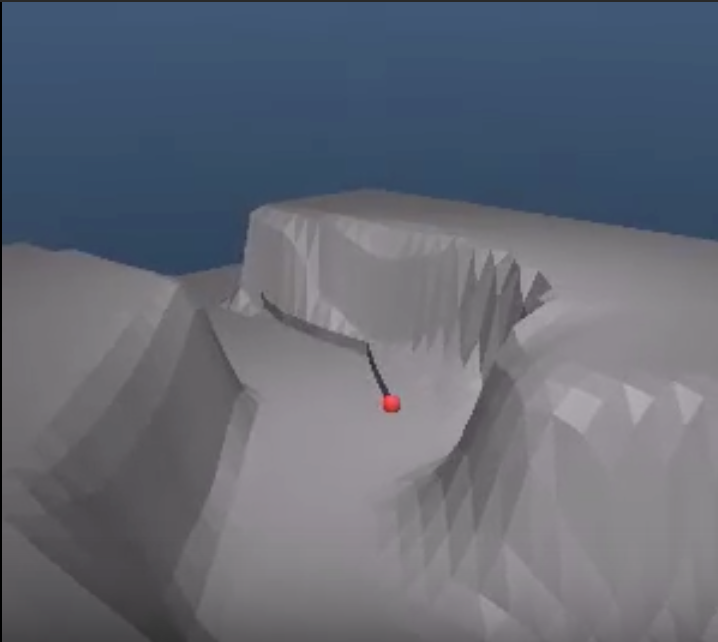}
\includegraphics[width=.2\textwidth]{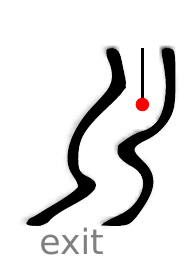}
\end{center}
\caption{\footnotesize
\emph{Left.} The target-seeking task. A top-down view of the arena showing the snake swimming to the green target region. \emph{Right.} The canyon traversal task. A snapshot of the snake in the maze and an illustration of the shape of the maze.
}
\label{fig:swimmerTasks}
\end{figure}

\textbf{Transfer task 1: Target-seeking} The first transfer task is a partially-observed target-seeking task with a sparse reward function (see Fig.\ \ref{fig:swimmerTasks}, left). The high-level controller egocentrically senses the vector from head to the green target region's center but only when the target center is within $\pm 60^{\circ}$ of the head direction. At the beginning of each episode, both the snake and target are deposited at random, with a minimum distance between them. To solve this task, the snake needs to learn a strategy to turn around until it sees the target and then swim towards it. Each episode lasts 800 time steps, and reward is delivered when the snake's head is within the target region. 

\textbf{Transfer task 2: Canyon traversal} The snake must swim through a simple canyon (Fig.\ \ref{fig:swimmerTasks}, right). Perceptual input to the high-level controller is given in the form of a $10$ pixel strip from an egocentric depth camera attached to the head. A positive reward is received when the swimmer has successfully navigated from start to end. An episode is terminated 25 steps after the snake's head has passed the canyon exit or after 3000 steps, whichever comes first.

\begin{figure}
\begin{center}
\begin{tabular}{c c}
\includegraphics[width=0.4\columnwidth]{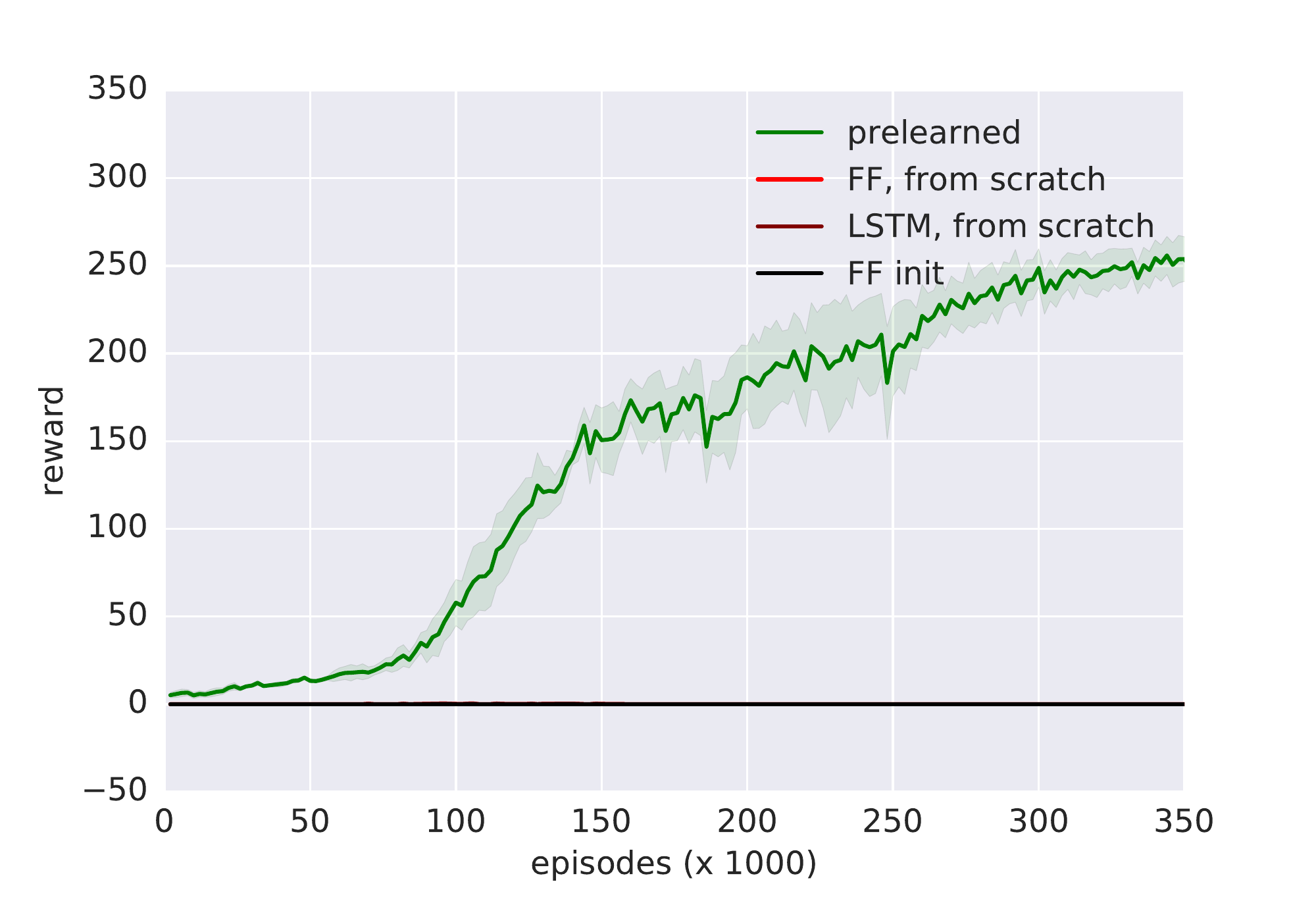} & 
\includegraphics[width=0.4\columnwidth]{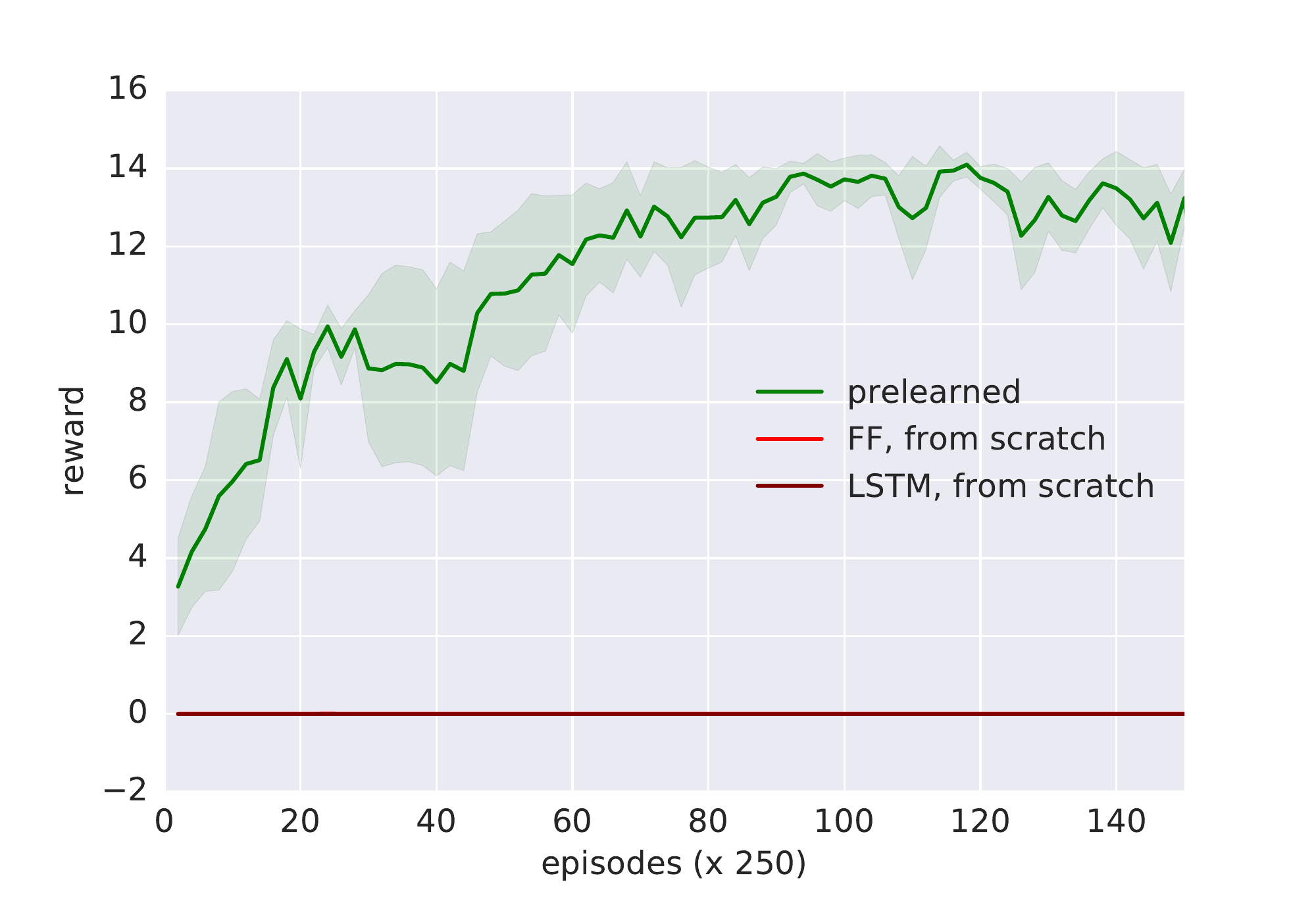}  

\\
{\small (a) target-seek } & {\small (b) canyon traversal}
\end{tabular}
\end{center}
\caption{\footnotesize
Performance of the pre-learned locomotor controller for the snake on: (a) the target-seeking task and (b) the canyon traversal problem.}
\label{fig:SwimmerTransfer} 
\end{figure}

Figure \ref{fig:SwimmerTransfer} shows the results of using the low-level controller to solve the transfer tasks. Both tasks are solved successfully when the learned motor primitives are harnessed. Without pre-training the low-level controller, however, an end-to-end system fails to learn. In both tasks, rewards are sparse: i.i.d.\ Gaussian exploration lacks the ability to make consistent progress in a given direction, so reaching the goals and receiving reward is highly unlikely; thus, no learning gradient exists.

\subsection{Quadruped}

\textbf{Pre-training task} Like the snake, the quadruped is initialized in a random orientation at the origin and must move to a target positioned randomly in its vicinity. The reward at each step is the (negative) distance to the target, and an episode lasts for 300 steps. The 34-dimensional input to the low-level controller is formed by the joint angles and angular velocities, the output of contact sensors attached to the legs and the ball-shaped torso, the velocity of the torso (translational and rotational) in the coordinate frame of the torso, and a 3-dimensional feature indicating the deviation of the torso's north pole from the z-axis. The high-level controller additionally receives the relative position of the target (36 dimensional input in total) and communicates with the low-level controller every 10 steps.

\begin{figure}
\begin{center}
\includegraphics[width=1.0\textwidth]{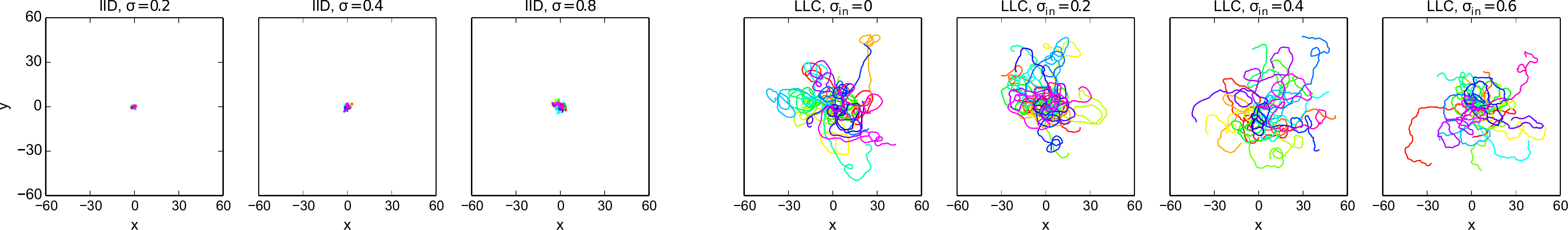}
\end{center}
\caption{\footnotesize
Random behavior of the quadruped when driven by i.i.d.\ random noise (left three plots) and using the pre-trained low-level controller (right four plots). Same format as in Figure \ref{fig:SwimmerNoise}
}
\label{fig:antNoise}
\end{figure}

\textbf{Analysis of the locomotor primitives} Figure \ref{fig:antNoise} provides an analysis of the behavior of the quadruped when actuated with i.i.d.\ Gaussian motor noise (left three plots) versus the randomly-modulated locomotor primitives (right four plots). Like the results shown for the snake in Figure \ref{fig:SwimmerNoise}, it is clear that the low-level controller achieves much better coverage of the space.

\begin{figure}
\begin{center}
\includegraphics[width=.36\textwidth]{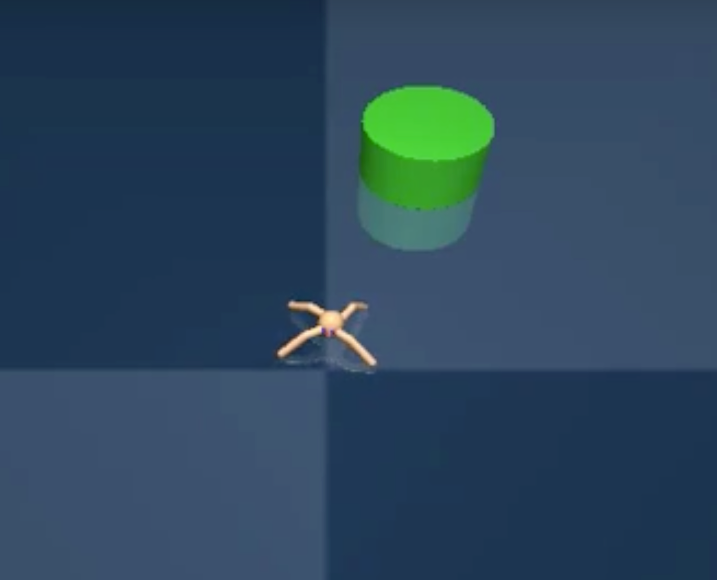}
\hspace{0.1\textwidth}
\includegraphics[width=.35\textwidth]{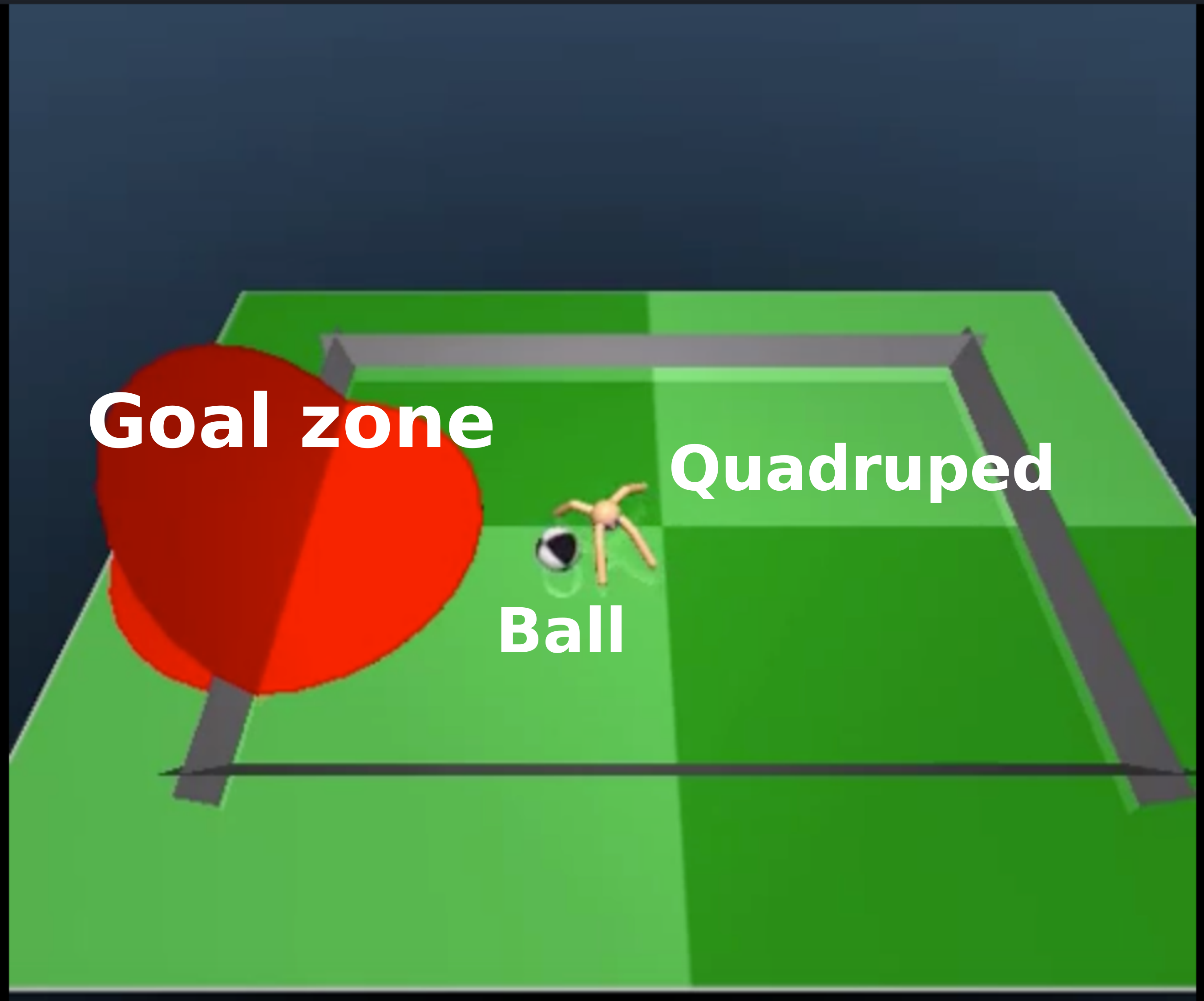}
\end{center}
\caption{\footnotesize
Transfer tasks for the quadruped: illustration of the target-seeking (left) and soccer task (right). }
\label{fig:antTasks}
\end{figure}

\textbf{Transfer task 1: Target-seeking} The quadruped has to move to a random target as quickly as possible.
Reward is given when the quadruped's torso is inside the green region. Episodes last for 600 steps. The difficulty of the task can be varied by adjusting the minimum initial distance between the quadruped and the target; if this distance is large, exploration becomes a key challenge. We evaluated the pre-learned low-level controller on two difficulty levels of the task, one with small minimum initial distances and one with large minimum distances, respectively. In both cases, the newly learned high-level controller receives the relative position of the target in addition to the proprioceptive features provided to the fixed low-level controller.

\textbf{Transfer task 2: Soccer} The quadruped is placed with a ball inside a walled pitch, as shown in Figure \ref{fig:antTasks}. To receive a reward, it needs to manipulate the ball into the red ``goal zone,'' which consists of two half-circles covering the floor and the back wall of the pitch. The goal zone has higher friction than the rest of the pitch to prevent the ball from simply bouncing away (simulating a goal net), and there is a small resistance that needs to be overcome when the ball is pushed into the zone while on the floor. 

We consider two versions of the task that differ in their initial conditions. In a restricted version the quadruped and ball are positioned and oriented randomly in non-overlapping areas close to the center line of the pitch. The initial position for the ball is closer to the goal than that of the quadruped. In the less restricted version both ball and quadruped can be initialized further away from the center line and the ball can be placed further away from the goal than the quadruped so that the quadruped has to move away from the goal to position itself behind the ball. Note that even in the restricted version the quadruped is not necessarily facing the goal after initialization.

The high-level controller senses the proprioceptive features alongside features indicating the relative displacements from both the ball and center of the goal and the velocity of the ball in the body frame. Reward is only provided when the ball has graced the goal zone. From this sparse feedback, the quadruped must learn both how to navigate to the ball and also how to manipulate the ball into the goal zone. The task poses a severe exploration problem to the quadruped as solutions of the two subtasks are not guided by shaping rewards.

\begin{figure}
\begin{tabular}{c c c}
\includegraphics[width=0.3\columnwidth]{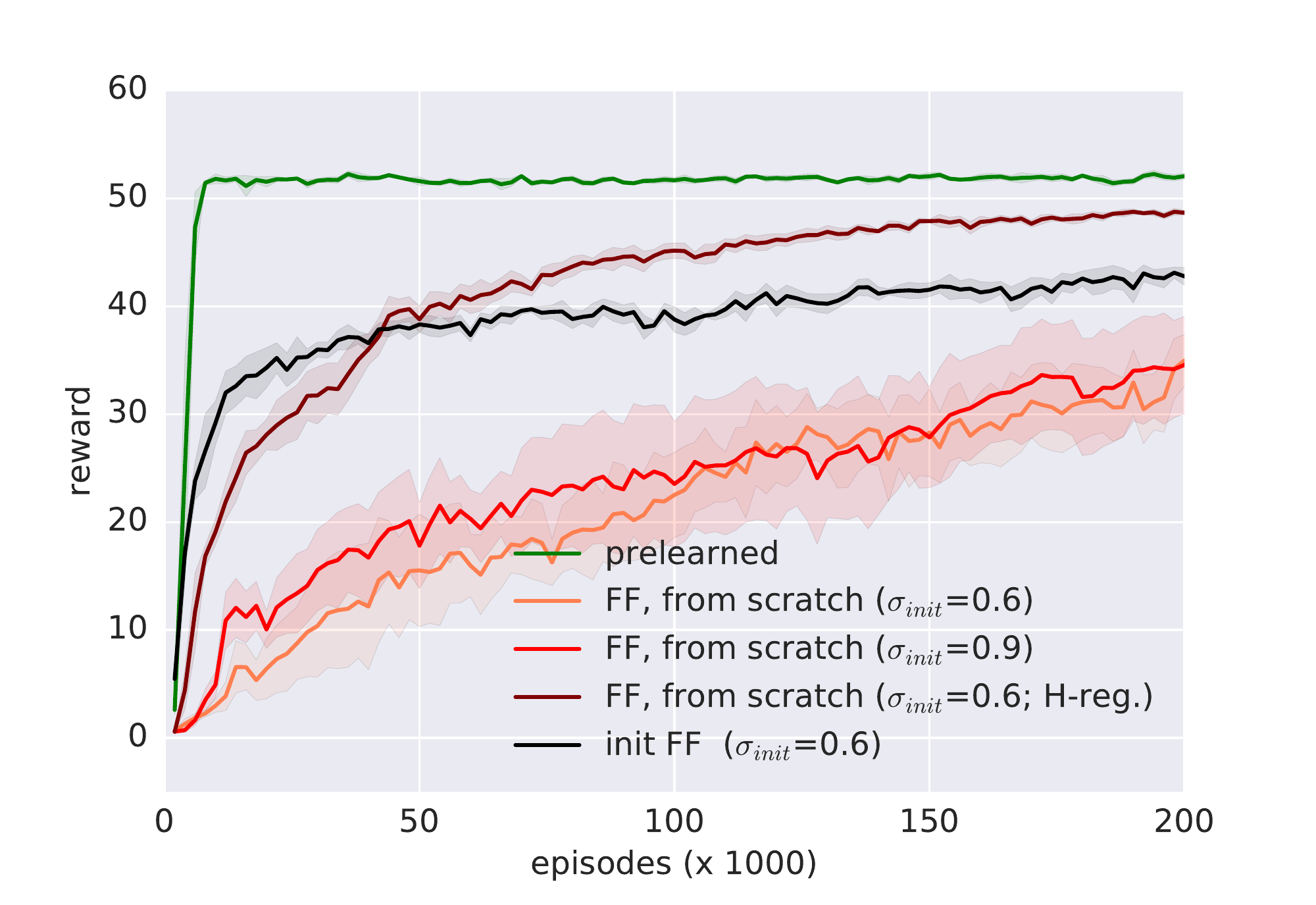} &
\includegraphics[width=0.3\columnwidth]{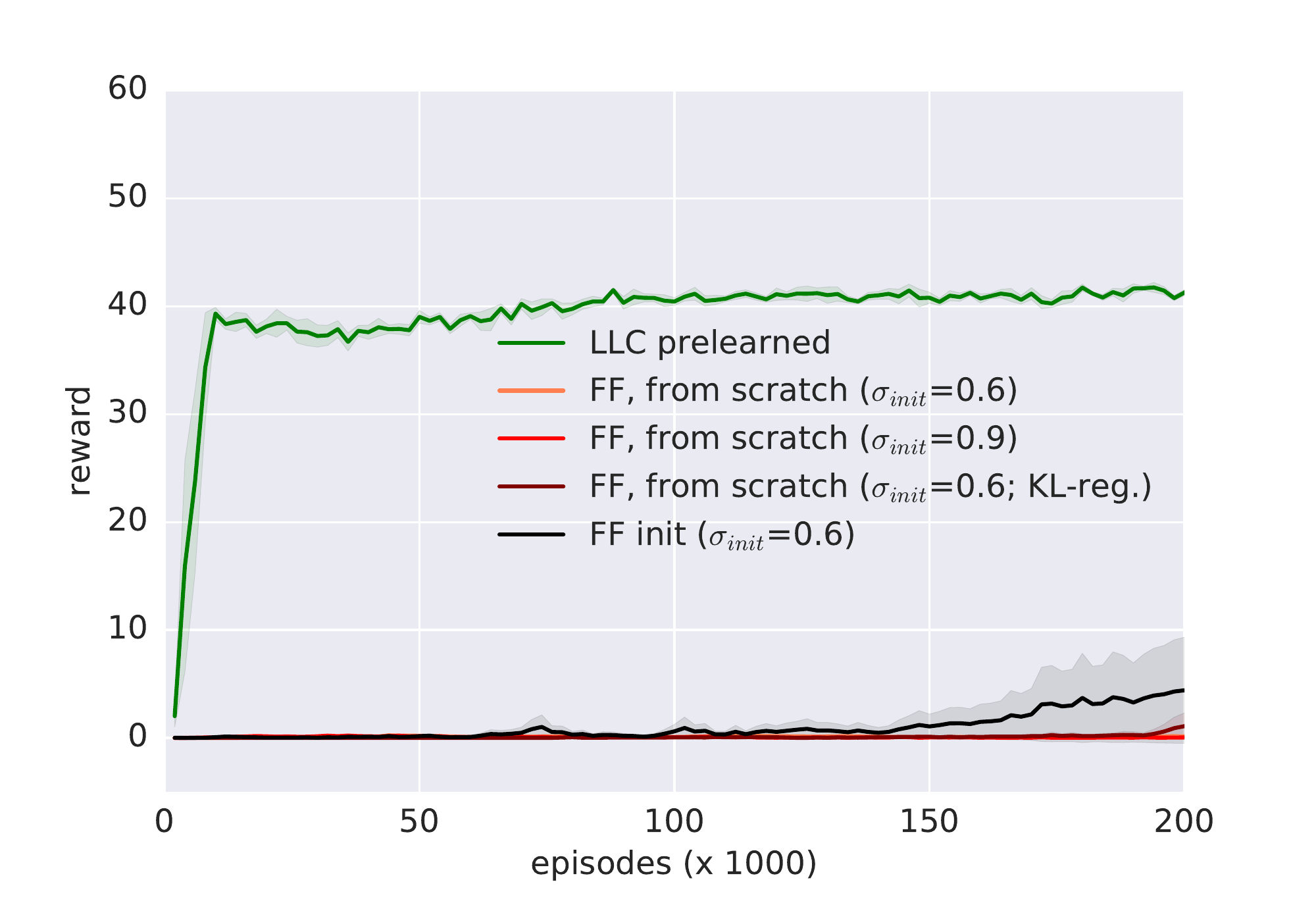} & 
\includegraphics[width=0.3\columnwidth]{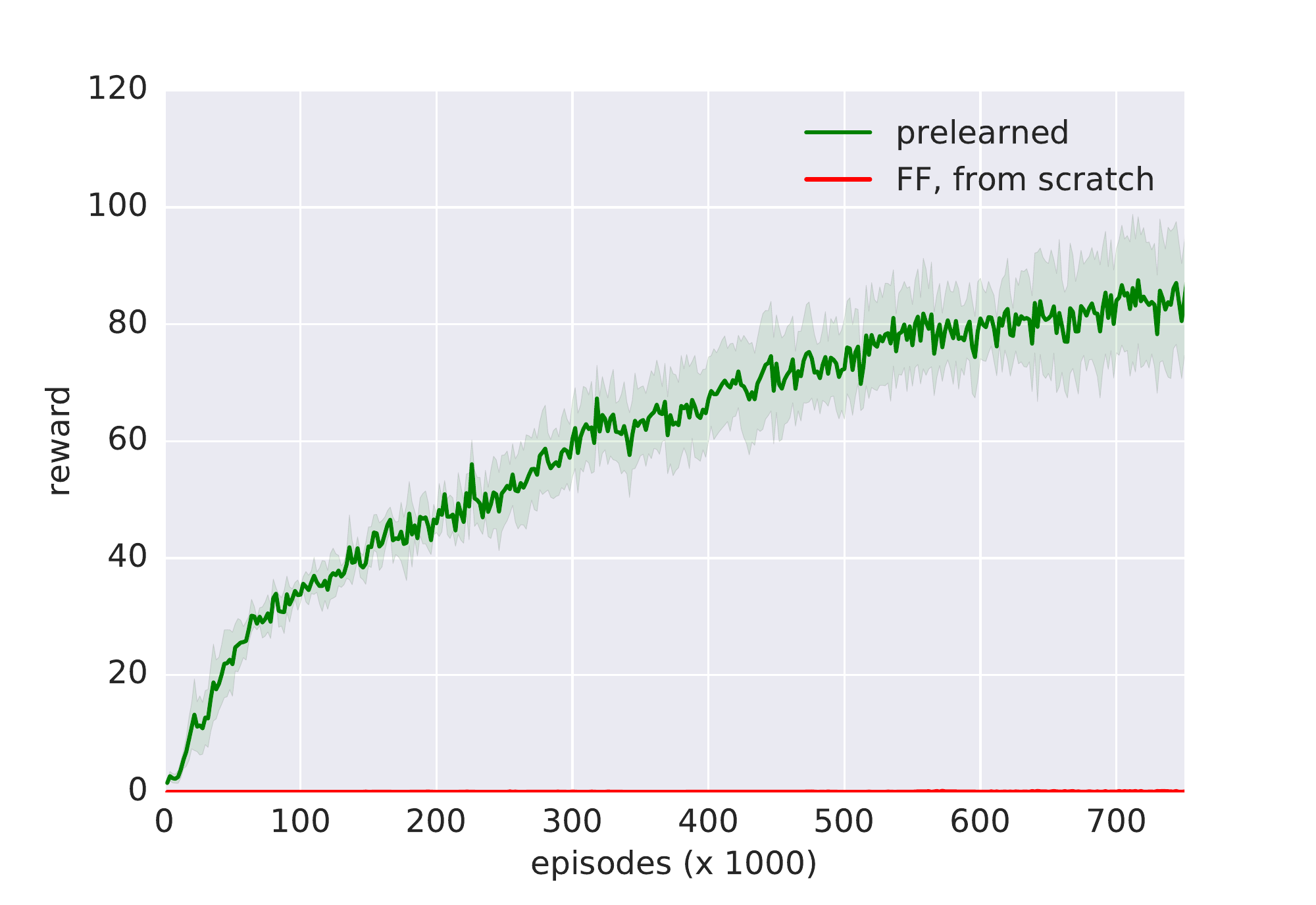}

\\
{\small (a) target-seek (easy)} & {\small (b) target-seek (hard) } & {\small (c) soccer}
\end{tabular}
\caption{\footnotesize
Performance of the pre-learned locomotor controller for the quadruped on the sparse reward go-to-target task ((a), (b); easy and hard, respectively), and the soccer problem (c). The results for learning from scratch with a FF network are sensitive to the initialization and regularization of the policy: Both, a larger initial value of the (learned) standard deviation ($\sigma_{init}$) as well as adding a regularizing term that encourages entropy in the per-step action distribution (as in \cite{mnih2016async}) improve the results (see (a)). The effectiveness of this approach is, however, noticeably reduced as task difficulty increases (see (b)). 
}
\label{fig:antTransfer}
\end{figure}

The results for the transfer tasks are shown in Figure \ref{fig:antTransfer}. 
For the target-seeking task, we show results for the two different levels of difficulty. The hierarchical policy with pre-trained locomotor primitives very rapidly learns to solve both task variants. In contrast, a policy that is trained from scratch only learns a satisfactory solution to the simpler version. Very little progress is made on the hard version of the task over the 200,000 episodes shown due to the very sparse learning signal and poor exploration. 

The hierarchical policy with locomotor primitives also makes good progress on the soccer task. For the restricted version of the task quantitative results are shown in Figure \ref{fig:antTransfer}c. We obtain players that learn to score goals for many initial configurations of the quadruped and ball. Attempts to solve the soccer task from scratch do not succeed. For the more challenging version learning is slower and results are more sensitive to initial conditions and learning parameters.  Nevertheless, we obtain several players that fetch the ball, pursue it if necessary, and score (see video).  

\subsection{Humanoid}

\begin{figure}
\begin{center}
\includegraphics[width=1\columnwidth]{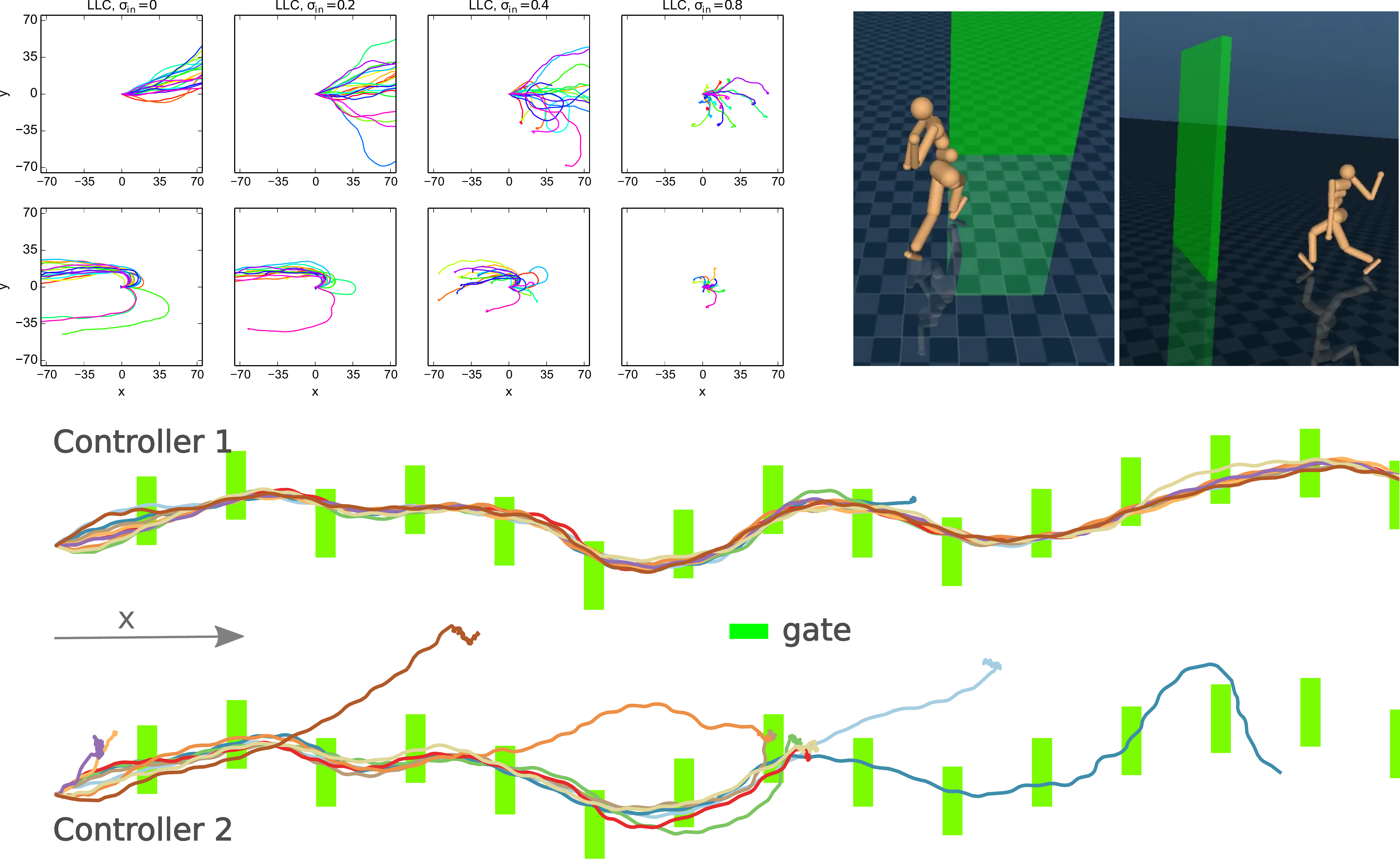}
\end{center}
\caption{\footnotesize
Humanoid: \emph{Top left}: Trajectories obtained by running two different low-level controllers in isolation. Stronger modulatory input increases the trajectory diversity but also increases the probability of falling. i.i.d.\ Gaussian exploration on the actions simply falls and makes no progress (see video). \emph{Top right}: Screen shot of the humanoid approaching a virtual gate in the transfer task (two perspectives). \emph{Bottom}: Trajectories through a series of gates obtained with different low-level controllers in the transfer task (10 trajectories with different initial conditions each). Not all learned  policies for the slalom task are equally successful (cf.\ controller 2). 
}
\label{fig:humanoidTask}
\end{figure}

In a final set of experiments, we applied the approach to a particularly challenging 27 degree-of-freedom control problem: the humanoid model shown in Figure \ref{fig:humanoidTask}. With 21 actuators the problem is much higher dimensional than the others. Moreover, whereas the snake and quadruped are passively stable, keeping the humanoid from falling is a non-trivial control problem in itself.

\textbf{Pre-training task} The training task consists of a simple multi-task setup: In every episode the humanoid is initialized in a standing position facing in the direction of the x-axis and is required to either move straight, or follow a leftward or a rightward circle of a fixed radius (5m). The reward function consists of a small quadratic control penalty, a positive constant stay-alive reward, the velocity in the desired direction (forward or along the circle) clipped at 2.5m/s, and, for the circle tasks, a quadratic penalty for deviating from the desired distance to the center point of the circle. Episodes last for up to 300 steps but are terminated when height of the humanoid falls below 0.9m.

The input to the low-level controller consists of proprioceptive features describing the configuration of the body relative to the model root (torso) and ground, and associated velocities. The high-level controller additionally receives the relative position of the target as input. The control interval for the high-level controller is $K=10$.

\textbf{Analysis of the locomotor primitives} Controlling the humanoid is a challenging problem and learning in the pre-training task is more sensitive to initial conditions and learning parameters than for the quadruped and snake. Nevertheless, we obtained several well performing policies with some variability across the gaits. Analyzing several of the associated low-level controllers in the same way as in the previous sections revealed that the locomotor primitives encode quite stable walking behaviors, typically taking hundreds of steps before falling (Fig.\ \ref{fig:humanoidTask}a).

\textbf{Transfer task: Slalom} Our transfer task consists of a slalom walk where the humanoid is presented with a sequence of virtual gates to pass through. A reward of 5 is given for passing a gate, and missing a gate results in termination of the episode. No other reward is given. After a gate has been passed, the next gate is positioned randomly to the left or the right of the previous one. The newly trained high-level controller receives the proprioceptive features provided to the low-level controller as input, as well as the relative position and orientation of the next gate.

We  trained high-level controllers to solve the transfer task using some of the pre-trained low-level controllers and obtained several good solutions in which the humanoid learned to actively navigate the gates. Nevertheless, as expected from the already somewhat diverse set of solutions to the pretraining task, not all low-level controllers were equally suitable for solving the transfer task. And for a given low-level controller we further observed a much stronger sensitivity to the learning parameters and the initial conditions during transfer (see also section \ref{sec:Appendix:Variability} in the appendix). Considering the complexity of the humanoid and the relatively small number of constraints imposed by the pre-training task this is, however, perhaps not too surprising. We expect that a richer, more constrained pre-training regime would lead to more uniformly versatile and robust low-level controllers.

\section{Related Work}
\label{sec:RelWork}
The notion that biological motor systems are hierarchical is ancient, dating to the 19th century \cite{hughlings1889comparative}. In the 20th century, Bernstein promulgated the notion of hierarchical control through ``motor synergies,'' or stereotyped, multi-joint muscle activation combinations \cite{bernstein1967co}. More recently, the notion of spinal motor primitives has been forwarded by Mussa-Ivaldi and Bizzi \cite{mussa2000motor} and others \cite{loeb1999hierarchical}. Motor primitives have resonated in robotics, especially as Dynamic Movement Primitives \cite{ijspeert2002learning}, which are low-dimensionality attractor systems that can simplify the learning of robot movements, and hierarchical robot control abstractions date to at least the 1980s \cite{brooks1986robust}. Modern control theorists have also considered abstract hierarchical architectures for manipulation \cite{todorov2005task} and more bio-mechanical descriptions of locomotion \cite{song2015neural}.  

The reinforcement learning literature has also explored a wide variety of temporal abstractions that wrap low-level control into \emph{options} \cite{sutton1999between} or \emph{skills} \cite{konidaris2009efficient}. These temporal abstractions may be applied to motor control \cite{da2014learning,thomas2012motor,wayne2014hierarchical}, may be transferred to new tasks \cite{andre2002state,asadi2007effective,ravindran2003relativized}, and may also incorporate information hiding principles \cite{dayan1993feudal,konidaris2009efficient}. However, 
this prior work
typically requires precise subgoals to be specified, and treats options or skills as atomic actions. 
In contrast, our low-level motor skills emerge organically from pre-training in the context of natural tasks; and our high-level controller modulates these skills in a flexible manner to achieve its ends. Recent work \cite{vezhnevets2016straw} proposes an architecture for discovering temporally extended macro actions from scratch.


\section{Conclusion}

We have provided a preliminary investigation of a hierarchical motor control architecture that can learn low-level motor behaviors and transfer them to new tasks. 
Our architecture contains two levels of abstraction that differ both in their access to sensory information and in the time scales at which they operate. 
Our design encourages the low-level controller to focus on the specifics of reactive motor control, while a high-level controller directs behavior towards the task goal by communicating a modulatory signal. 

Our investigation departs from the common but unnatural setting where an agent is trained on a single task. Instead, we exploit the fact that many complex motor behaviors share low-level structure by building reusable low-level controllers for a variety of tasks.
We found our method to be especially effective when attempting challenging transfer tasks with sparse rewards where exploration via ``motor babbling'' is unlikely to accrue reward at all. This is illustrated in the transfer tasks for the swimmer, quadruped, and humanoid, in which direct end-to-end learning failed, but our method produced solutions.

We believe that the general idea of reusing learned
behavioral primitives is important, and the design principles we have followed represent possible steps towards this goal. 
Our hierarchical design with information hiding has enabled the construction of low-level motor behaviors that are sheltered from task-specific information, enabling their reuse. However, the detailed individual and joint contributions of the features of our architecture remain to be investigated more thoroughly in future work (in particular the role and relevance of different time scales), alongside strategies to increase the reliability and stereotypy of the low-level behaviors, especially for difficult control problems such as humanoid walking. 

Our approach currently depends on the assumption that we can propose a set of low-level tasks that are simple to solve yet whose mastery in turn facilitates the solution of other high-level tasks. 
For motor behavior, we believe this assumption holds rather generally, as walking, for example, underpins a multitude of more complex tasks.
By instilling a relatively small number of reusable skills into the low-level controllers, we expect that a large number of more complicated tasks involving composition of multiple behaviors should become solvable. We believe that this general direction could open new avenues for solving complex real-world, robotic control problems.

\paragraph{Acknowledgements}

We would like to thank Tom Schaul, Tom Erez, Ziyu Wang, Sasha Vezhnevets, and many others of the DeepMind team for helpful discussions and feedback.

\bibliographystyle{plain}
{\footnotesize 
\linespread{0} \bibliography{references}

\begin{thebibliography}{10}

\bibitem{andre2002state}
D~Andre and S~J Russell.
\newblock State abstraction for programmable reinforcement learning agents.
\newblock In {\em AAAI/IAAI}, pages 119--125, 2002.

\bibitem{asadi2007effective}
M~Asadi and M~Huber.
\newblock Effective control knowledge transfer through learning skill and
  representation hierarchies.
\newblock In {\em IJCAI}, volume~7, pages 2054--2059, 2007.

\bibitem{bernstein1967co}
N~A Bernstein.
\newblock {\em The co-ordination and regulation of movements}.
\newblock Pergamon Press Ltd., 1967.

\bibitem{brooks1986robust}
R~A Brooks.
\newblock A robust layered control system for a mobile robot.
\newblock {\em Robotics and Automation, IEEE Journal of}, 2(1):14--23, 1986.

\bibitem{da2014learning}
B~C Da~Silva, G~Baldassarre, G~Konidaris, and A~Barto.
\newblock Learning parameterized motor skills on a humanoid robot.
\newblock In {\em Robotics and Automation (ICRA), 2014 IEEE International
  Conference on}, pages 5239--5244. IEEE, 2014.

\bibitem{dayan1993feudal}
P~Dayan and G~E Hinton.
\newblock Feudal reinforcement learning.
\newblock In {\em Advances in neural information processing systems}, pages
  271--271. Morgan Kaufmann Publishers, 1993.

\bibitem{dominici2011locomotor}
N~Dominici, Y~P Ivanenko, G~Cappellini, A~d'Avella, V~Mond{\`\i}, M~Cicchese,
  A~Fabiano, T~Silei, A~Di~Paolo, C~Giannini, et~al.
\newblock Locomotor primitives in newborn babies and their development.
\newblock {\em Science}, 334(6058):997--999, 2011.

\bibitem{gu2016continuous}
S~Gu, T~P Lillicrap, I~Sutskever, and S~Levine.
\newblock Continuous deep q-learning with model-based acceleration.
\newblock {\em arXiv preprint arXiv:1603.00748}, 2016.

\bibitem{heess2015learning}
N~Heess, G~Wayne, D~Silver, T~P Lillicrap, T~Erez, and Y~Tassa.
\newblock Learning continuous control policies by stochastic value gradients.
\newblock In {\em Advances in Neural Information Processing Systems}, pages
  2926--2934, 2015.

\bibitem{hughlings1889comparative}
J~Hughlings~Jackson.
\newblock On the comparative study of disease of the nervous system.
\newblock {\em Br Med J}, 17:355--362, 1889.

\bibitem{ijspeert2002learning}
A~J Ijspeert, J~Nakanishi, and S~Schaal.
\newblock Learning attractor landscapes for learning motor primitives.
\newblock In {\em Advances in Neural Information Processing Systems 15}, 2002.

\bibitem{kandel2000principles}
E~R Kandel, J~H Schwartz, T~M Jessell, et~al.
\newblock {\em Principles of neural science}, volume~4.
\newblock McGraw-hill New York, 2000.

\bibitem{kingma2013}
D~P Kingma and M~Welling.
\newblock {Auto-encoding variational Bayes}.
\newblock {\em arXiv preprint arXiv:1312.6114}, 2013.

\bibitem{konidaris2009efficient}
G~Konidaris and A~G Barto.
\newblock Efficient skill learning using abstraction selection.
\newblock In {\em IJCAI}, volume~9, pages 1107--1112, 2009.

\bibitem{levine2015end}
S~Levine, C~Finn, T~Darrell, and P~Abbeel.
\newblock End-to-end training of deep visuomotor policies.
\newblock {\em arXiv preprint arXiv:1504.00702}, 2015.

\bibitem{lillicrap2015continuous}
T~P Lillicrap, J~J Hunt, A~Pritzel, N~Heess, T~Erez, Y~Tassa, D~Silver, and
  D~Wierstra.
\newblock Continuous control with deep reinforcement learning.
\newblock {\em arXiv preprint arXiv:1509.02971}, 2015.

\bibitem{loeb1999hierarchical}
G~E Loeb, I~E Brown, and E~J Cheng.
\newblock A hierarchical foundation for models of sensorimotor control.
\newblock {\em Experimental brain research}, 126(1):1--18, 1999.

\bibitem{mnih2016async}
V~Mnih, A~P Badia, M~Mirza, A~Graves, T~P Lillicrap, T~Harley, D~Silver, and
  K~Kavukcuoglu.
\newblock Asynchronous methods for deep reinforcement learning.
\newblock {\em CoRR}, abs/1602.01783, 2016.

\bibitem{mussa2000motor}
F~A Mussa-Ivaldi and E~Bizzi.
\newblock Motor learning through the combination of primitives.
\newblock {\em Philosophical Transactions of the Royal Society of London B:
  Biological Sciences}, 355(1404):1755--1769, 2000.

\bibitem{peiper1929schreitbewegungen}
A~Peiper.
\newblock Die {S}chreitbewegungen der {N}eugeborenen [{T}he walking movements
  of newborns].
\newblock {\em Monatsschrift fur Kinderheilkunde}, 45:444, 1929.

\bibitem{ravindran2003relativized}
B~Ravindran and A~G Barto.
\newblock Relativized options: Choosing the right transformation.
\newblock In {\em ICML}, pages 608--615, 2003.

\bibitem{rezende2014}
D~Rezende, S~Mohamed, and D~Wierstra.
\newblock {Stochastic Backpropagation and Approximate Inference in Deep
  Generative Models}.
\newblock In Tony Jebara and Eric~P. Xing, editors, {\em International
  Conference on Machine Learning}, 2014.

\bibitem{schulman2015high}
J~Schulman, P~Moritz, S~Levine, M~Jordan, and P~Abbeel.
\newblock High-dimensional continuous control using generalized advantage
  estimation.
\newblock {\em arXiv preprint arXiv:1506.02438}, 2015.

\bibitem{song2015neural}
S~Song and H~Geyer.
\newblock A neural circuitry that emphasizes spinal feedback generates diverse
  behaviours of human locomotion.
\newblock {\em The Journal of physiology}, 593(16):3493--3511, 2015.

\bibitem{sutton1999between}
R~S Sutton, D~Precup, and S~Singh.
\newblock Between mdps and semi-mdps: A framework for temporal abstraction in
  reinforcement learning.
\newblock {\em Artificial intelligence}, 112(1):181--211, 1999.

\bibitem{thomas2012motor}
P~S Thomas and A~G Barto.
\newblock Motor primitive discovery.
\newblock In {\em ICDL-EPIROB}, pages 1--8, 2012.

\bibitem{todorov2005task}
E~Todorov, W~Li, and X~Pan.
\newblock From task parameters to motor synergies: A hierarchical framework for
  approximately optimal control of redundant manipulators.
\newblock {\em Journal of robotic systems}, 22(11):691--710, 2005.

\bibitem{vezhnevets2016straw}
Alexander Vezhnevets, Volodymyr Mnih, John Agapiou, Simon Osindero, Alex
  Graves, Oriol Vinyals, and Koray Kavukcuoglu.
\newblock Strategic attentive writer for learning macro-actions.
\newblock {\em CoRR}, abs/1606.04695, 2016.

\bibitem{wayne2014hierarchical}
G~Wayne and LF~Abbott.
\newblock Hierarchical control using networks trained with higher-level forward
  models.
\newblock {\em Neural computation}, 2014.

\end{thebibliography}

\clearpage
\appendix

\section{Network Parameters}

We use the same hierarchical neural network architecture for all problems and only change the number of hidden units. The outputs of both the high- and low-level parameterize means and standard deviations. The standard deviations are produced by a linear layer with sigmoidal output nonlinearity.

The low-level controller is a standard feedforward network with three hidden layers using $\tanh()$ nonlinearities. The second layer outputs are concatenated with the output of the high-level controller. The high-level controller only outputs 10 units as a bottleneck. For the snake and quadruped, there are 150 hidden units per layer aside from the concatenation layer; for the humanoid 300 hidden units.

During the pre-training the high-level controller is an LSTM with a perceptual encoder. The perceptual encoder has 30, 40, and 100 hidden units with $\tanh()$ nonlinearity for the snake, quadruped, and humanoid. The LSTM has 10 cells. Note that there is a separate high-level controller for the three humanoid subtasks (run straight, left circle, right circle).

For transfer the high-level controller LSTM has 50 cells for the two snake tasks and a perceptual encoder with 100 hidden units. For the quadruped seek-target task we used the same configuration as during pre-training; but a LSTM with 50 cells and a perceptual encoder with 100 hidden units for the soccer task. The high-level controller for the humanoid slalom task has 50 LSTM cells and an encoder with 300 hidden units. 

The non-hierarchical control experiments were run with feedforward networks of 2 layers of 300 hidden units for all experiments. 

\section{Experimental procedure for transfer tasks}

\paragraph{Pre-training}
To obtain a low-level controller we trained a full policy network as shown schematically in Fig.\ 1 (main text) on a suitable pre-training task. The pre-training task was chosen to be easy to solve but also to require a sufficiently flexible low-level controller. For the snake and the quadruped these were simple go-to-target tasks in which the respective agent needed to navigate to a random target, requiring both forward movement and turning. A smooth shaping reward was used during pre-training.

For the humanoid we experimented with several pre-training tasks. The results in the main text were obtained by training a single low-level controller in a simple multi-task setup: walk straight, walk in a left circle with a fixed radius, walk in a right circle with a fixed radius. The task was sampled randomly and uniformly at the beginning of each episode. The low-level controller was shared across the three tasks, but we used a separate high-level controller and value function for each of them.

For the results in the paper the high-level controller did not emit a stochastic but a deterministic control signal to the low-level controller during pre-training.

\paragraph{Transfer}
After pre-training we removed the part of the network corresponding to the high-level controller and replaced it with either IID Gaussian noise (e.g.\ for the plots in Fig.\ 3,6 in the main text), or with a randomly initialized \emph{stochastic} network for the transfer tasks. It is worth noting that the high-level controller used for transfer did not have the same structure as the one used for pre-training. Besides being stochastic it had typically more capacity, and it operated on a different observation space (due to the change in task). 

The parameters of the pre-learned low-level controller were frozen during transfer. However, for low-level controllers that learned the standard-deviation of the low-level sampling distribution, we re-initialized and re-learned also this final linear output layer that produced the pre-sigmoid standard-deviation.

We performed a grid-search over hyper-parameters both for the pre-training and transfer experiments (see main text). We generally found the transfer experiments to be more sensitive to the choice of hyper-parameters and initial conditions, possibly due to the increased task difficulty.

\section{Additional analyses}

\subsection{Variability across low-level controllers}
\label{sec:Appendix:Variability}
The parameter sweep performed during pre-training leaves us with several candidate policies from which to extract low-level controller for the transfer task. In the main paper we present transfer results for a particular low-level controller for quadruped and swimmer respectively, which we picked manually but at random among the well-performing policies obtained during pre-training (according to the average reward on the pre-training task). 

Below we show transfer results for several other low-level controllers for the snake, the quadruped, and the humanoid which were extracted from different policies which had achieved similar performance on the pre-training task. For each such low-level controller we performed a sweep over hyperparameters for learning a new high-level controller for solving the target-seeking transfer task. In Fig.\ \ref{fig:appendix:coreComparison} we show averages of the five best performing controllers obtained from each sweep. (Note that this deviates from the paradigm in the main text where we show an average over runs for the best performing setting of the hyper-parameters.)

\begin{figure}[h!]
\begin{center}
\begin{tabular}{c c c}
\includegraphics[width=.3\textwidth]{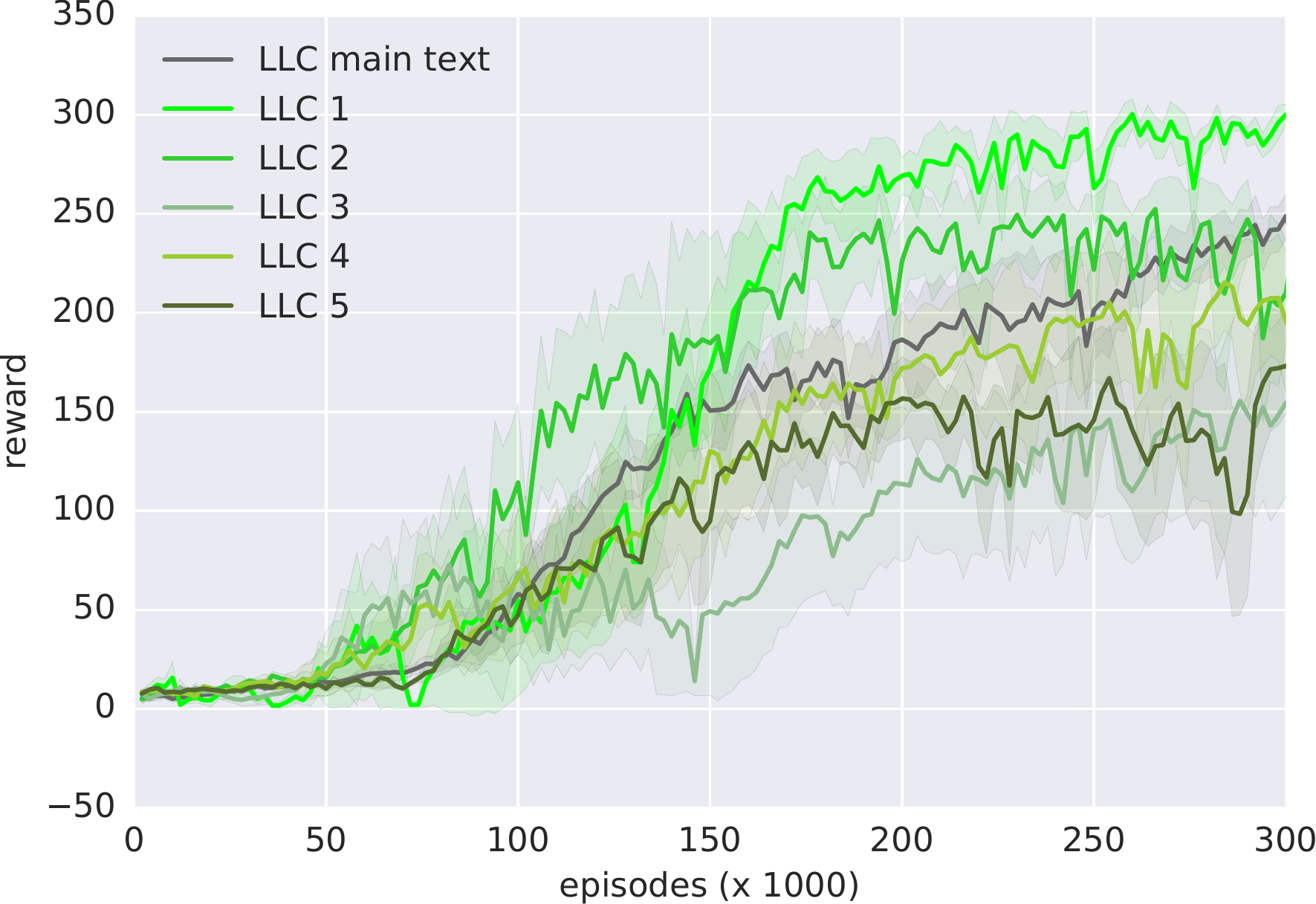} & 
\includegraphics[width=.3\textwidth]{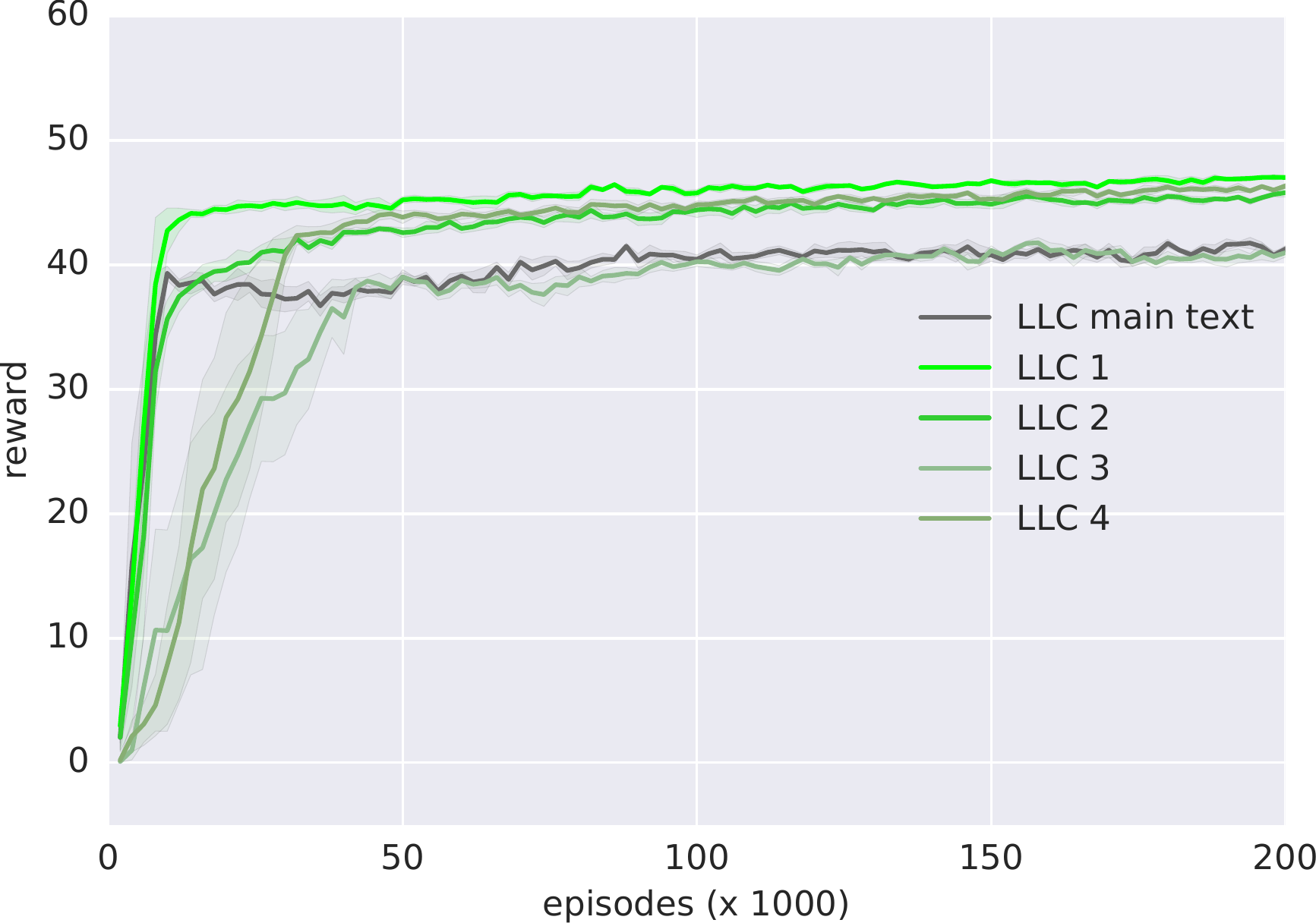} &
\includegraphics[width=.3\textwidth]{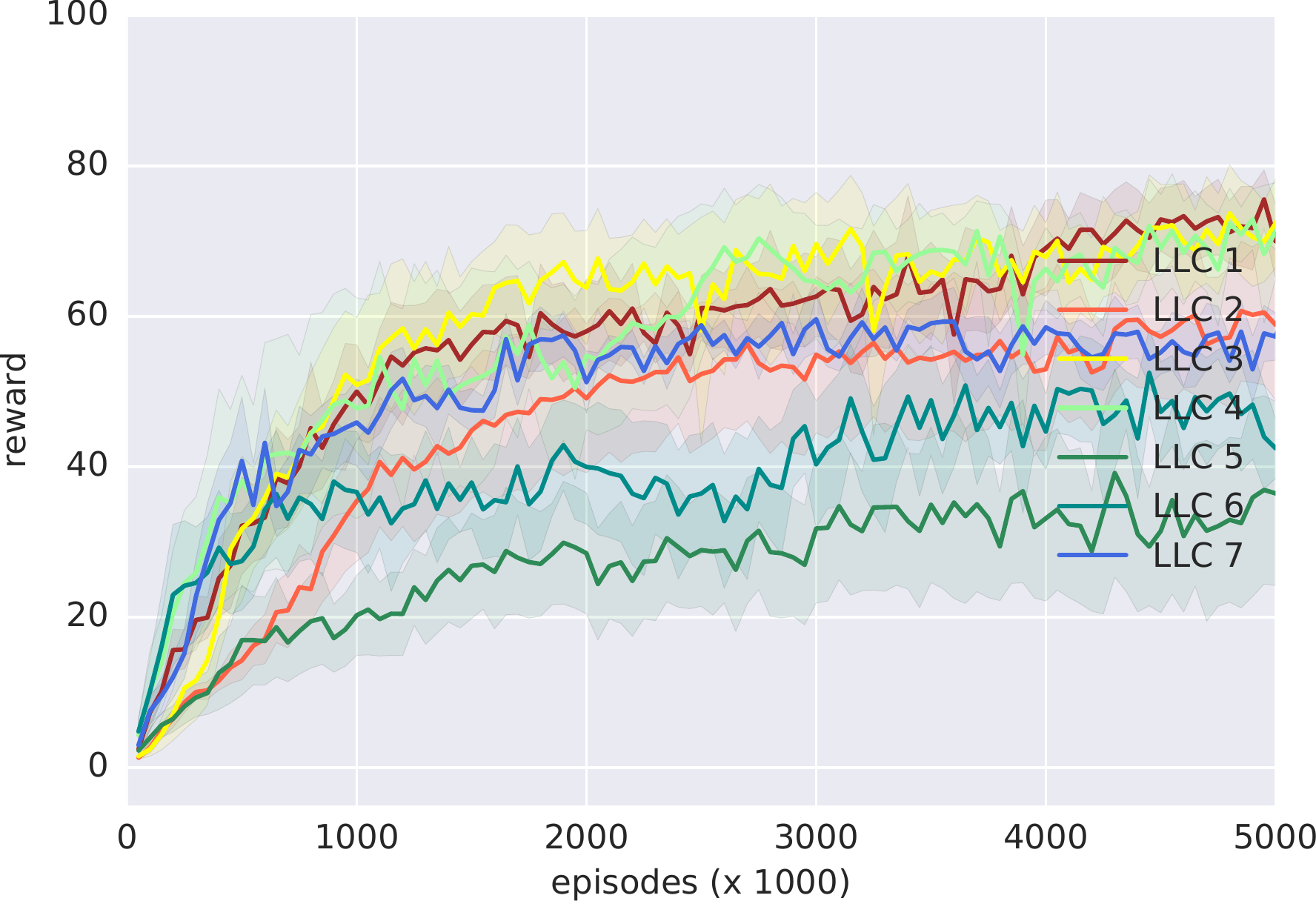} \vspace{0.2cm}
\\
{\small (a) Snake } & {\small (b) Quadruped } & {\small (c) Humanoid }
\end{tabular}
\end{center}
\caption{\footnotesize
Performance of several policies obtained with different pre-trained low-level controllers on the target-seeking transfer tasks for \emph{(a)} the snake, and \emph{(b)} the quadruped. We also show \emph{(c)} the performance of several low-level controllers for the humanoid on the slalom task. Each trace shows an average over the five best performing controllers obtained from the parameter sweep. For snake and quadruped the controllers used in the main text (Figs.\ 4a and 7b) are shown in dark gray.
}
\label{fig:appendix:coreComparison}
\end{figure}

While there is clearly some variability across the snake and quadruped low-level controllers all of them led to at least some reasonable solutions for the transfer task. For the humanoid the variability was larger both across pre-learned low-level controllers and also for a particular low-level controller with respect to the reliability across seeds and different random initializations of the high-level policy: For some of the pre-learned humanoid low-level controllers we obtain only a small number of robust solutions to the slalom task. (Note that a reward of 5 was provided every time the humanoid crossed a gate. Thus, e.g.\ an average reward of 30 means that the agent passes six gates on average per episode. The supplemental video shows some of the diversity we observed.) Poor performing slalom controllers seem to often fail at the beginning of a new episode due to a lack of robustness to the random initial states.

\subsection{Learned low-level behavior and high-level exploration}

We further investigated the effect of some of the architectural choices that we made in the experiments in the main text, in particular the effect of the high-level noise, and the effect of the high-level control interval. The results of this analysis are shown in Fig.\ \ref{fig:appendix:parameterComparison} for multiple low-level controllers for each of the three creatures. 

As in the main text we show trajectories resulting from replacing the high-level control signal from a learned high-level controller with Gaussian noise for several snake and quadruped low-level controller. 

This setup corresponds roughly to the situation that is encountered early in a transfer experiment (when the distribution modeled by the the high-level controller is roughly zero mean with a standard-deviation set to an initial value). Holding the noise fix for 10 steps corresponds to the high-level control interval used for the experiments in the main text. 

As in the main text (Figs.\ 3 and 6) the snake and quadruped are initialized in a random configuration in the center of plot at the beginning of each trajectory. The humanoid is always initialized facing in the direction of the  Also, we set the standard deviation of the low-level action distribution to 0.3 as at the beginning of pre-training. (Note that the random behavior of the low-level controller is the consequence of noise injected both at the primitive action level and via the high-level control.)

\begin{figure}[h!]
\begin{center}
\begin{tabular}{c c}
{\small \bf K=10 } & {\small \bf K=1 } \vspace{0.5cm}\\
\includegraphics[width=.45\textwidth]{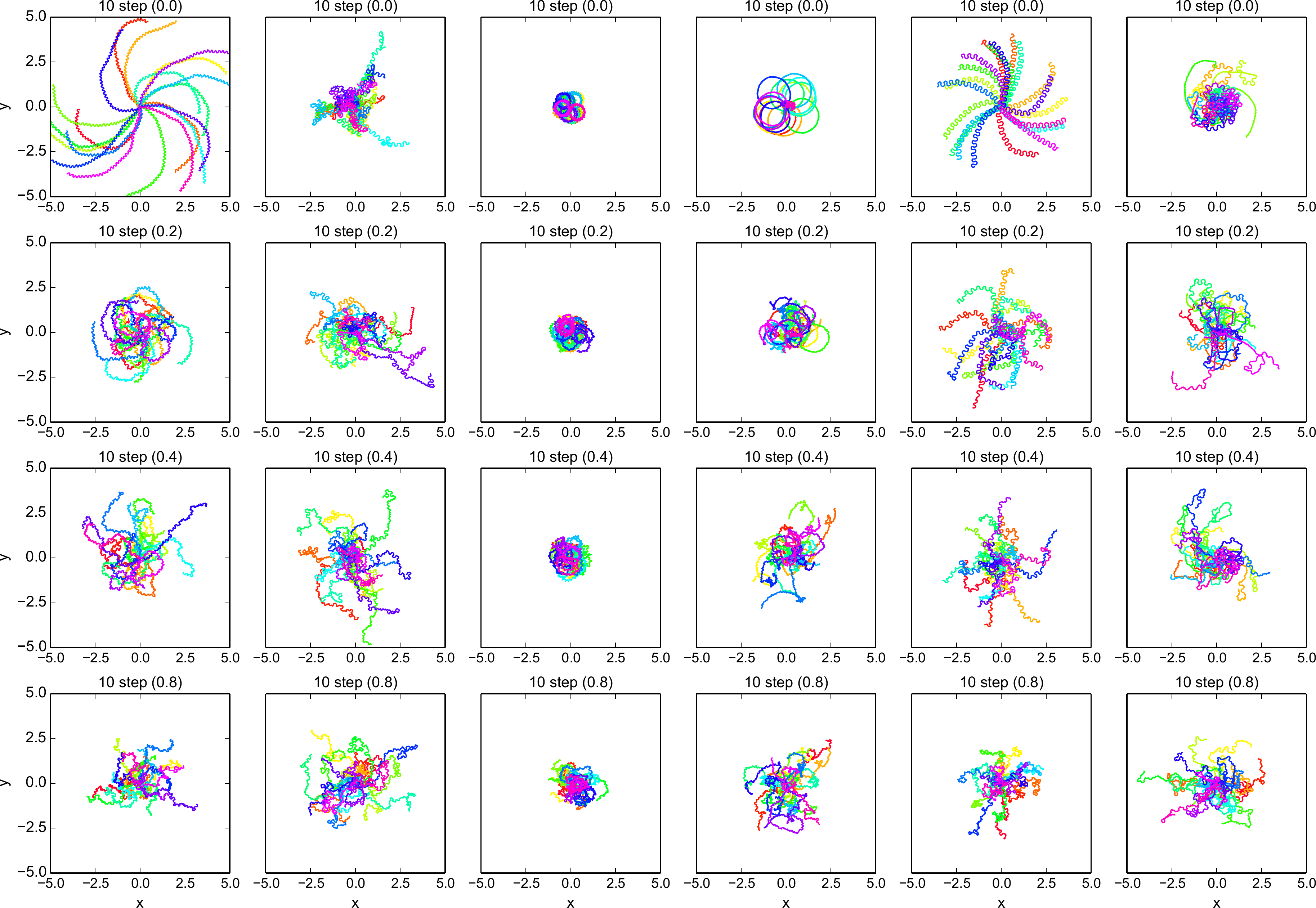} & 
\includegraphics[width=.45\textwidth]{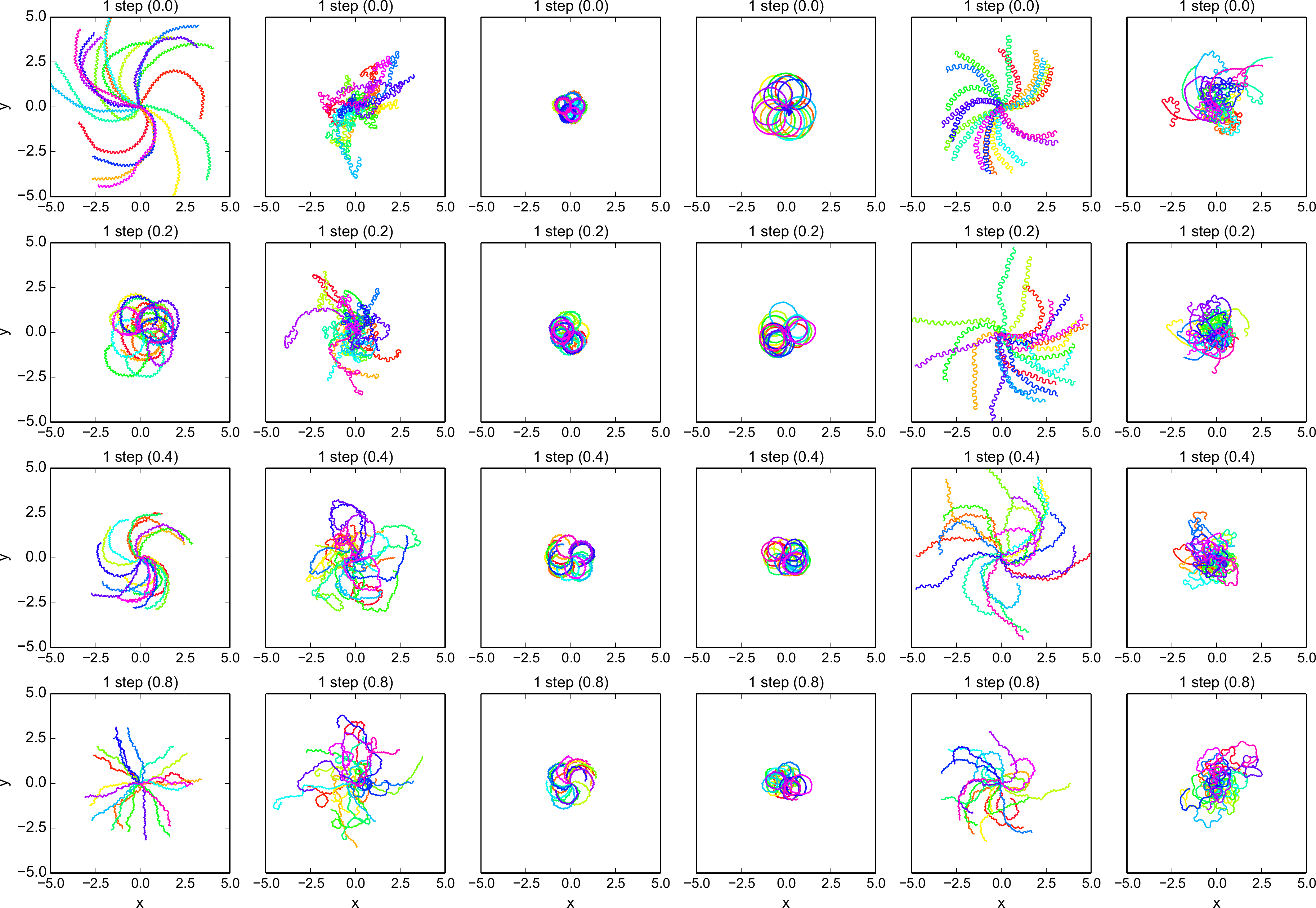} \vspace{0.1cm} \\
{\small (a)  Snake, K=10} & {\small (b) Snake, K=1 } \vspace{0.5cm}\\
\includegraphics[width=.45\textwidth]{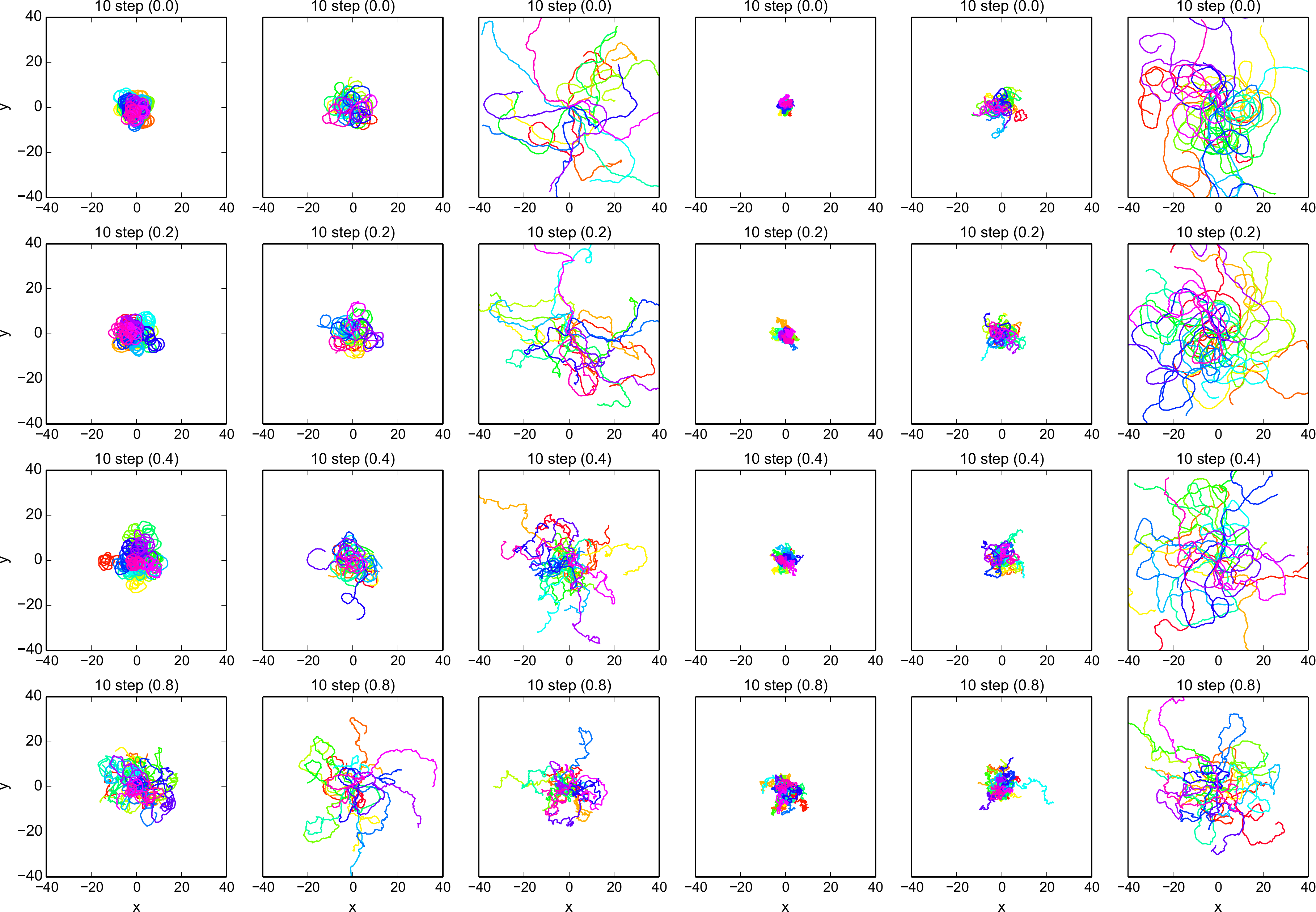} &
\includegraphics[width=.45\textwidth]{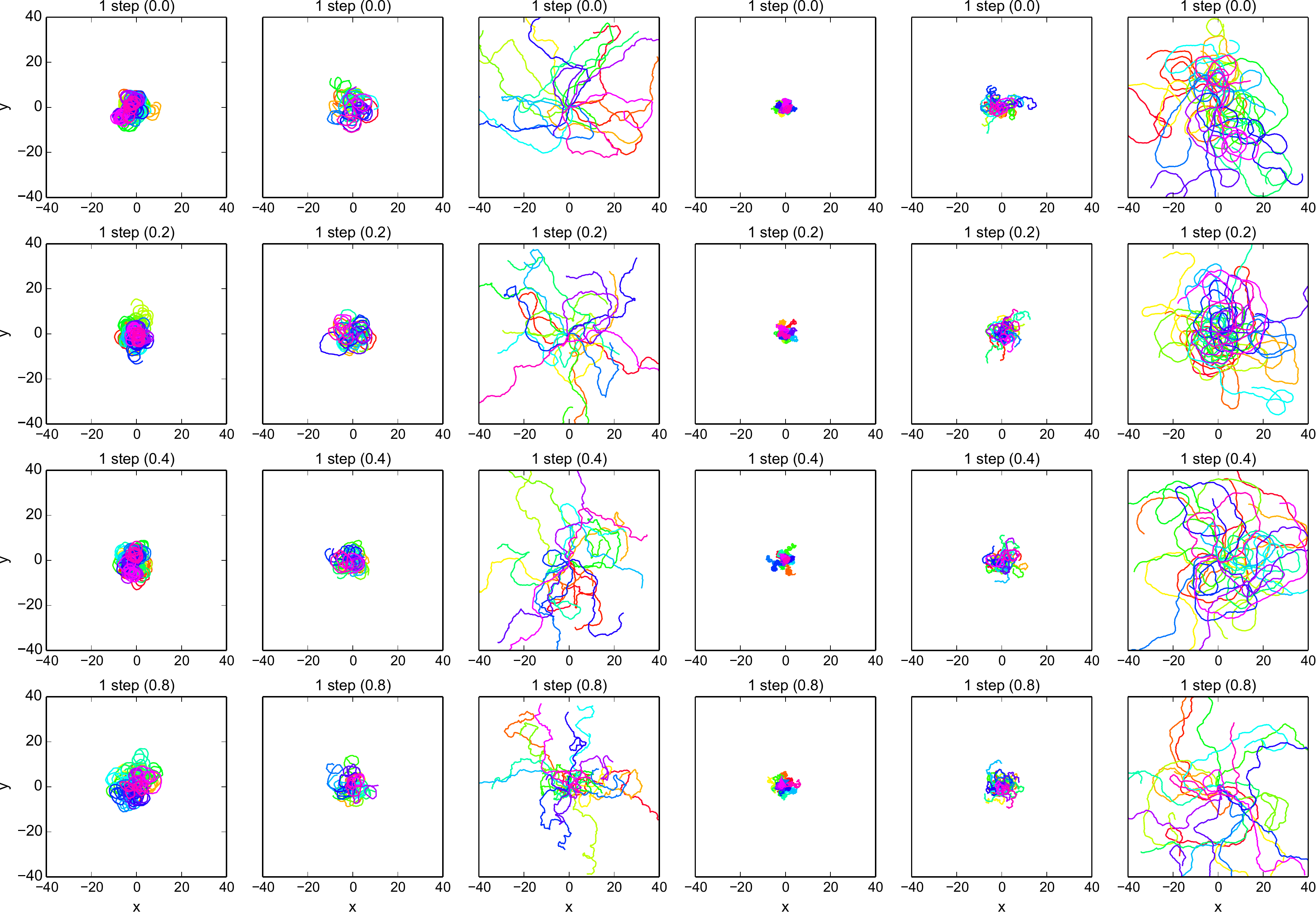} \vspace{0.1cm}\\
{\small (c) Quadruped, K=10} & {\small (d) Quadruped, K=1} \vspace{0.5cm}\\
\includegraphics[width=.45\textwidth]{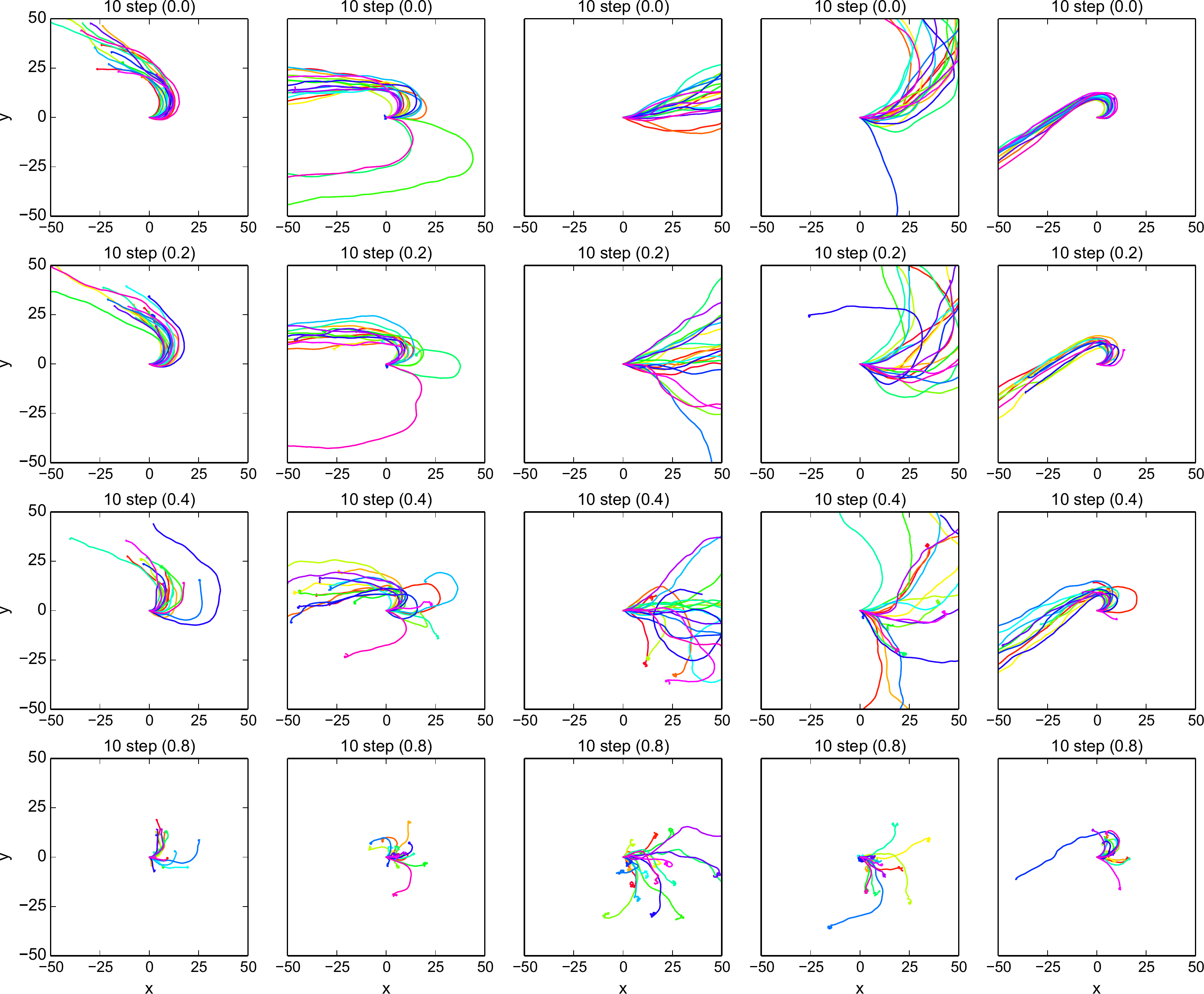} &
\includegraphics[width=.45\textwidth]{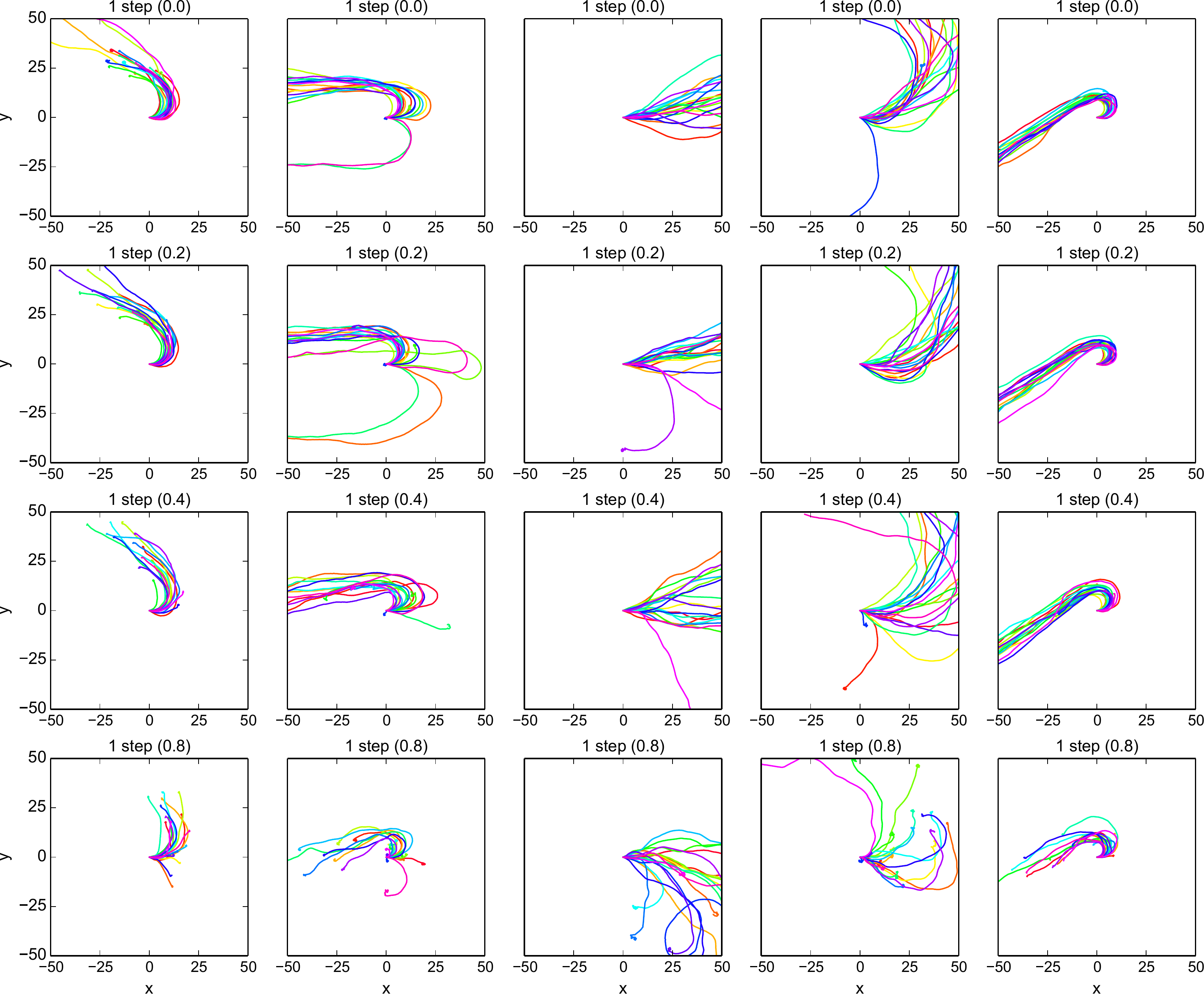} 
\vspace{0.1cm}\\
{\small (e) Humanoid, K=10} & {\small (f)  Humanoid, K=1} 
\end{tabular}
\end{center}
\caption{\footnotesize
Random trajectories obtained with pre-trained low-level controllers for snake (top;a,b), quadruped (middle; c,d), and humanoid (bottom; e,f) modulated by a high-level control signals drawn IID from a Gaussian  distributions with different variances (std.-dev. 0.0 to 0.8; see plots). The high-level control signal is held constant for 10 steps (left column of plots; a,c,e) or 1 step (right column of plots; b,d,f). Within each matrix of plots every column corresponds to a different pre-learned low-level controller, and the std-dev.\ of the high-level control signal varies across rows. The trajectories show the evoluion of the $x,y$ coordinates of the root joint of the creature. The trajectory length is 4000 steps for snake and quadruped and 2000 for humanoid controllers. For the former two the body is initialized at (0,0) in a random configuration (i.e.\ facing in a random direction) at the beginning of each trajectory. The humanoid initially faces in the direction of the x-axis.
}
\label{fig:appendix:parameterComparison}
\end{figure}

Several observations can be made:
Firstly, we obtain a diversity of behaviors across different low-level controllers for each creature.
Secondly, without any high-level control noise (top row) the behavior of snake and humanoid controllers is rather stereotyped, e.g.\ swimming in almost straight lines (left-most snake controller) or in circles of fixed radius (snake controller in fourth column). Note that for snake and quadruped the randomized initial configuration induces some variation across trajectories even in this case.
Thirdly, different controllers are affected differently by the high-level modulation but in general the trajectories of the snake and the humanoid become more diverse as the modulation strength increases (although the humanoid falls easily for large std.-devs.: see bottom row of Fig.\ \ref{fig:appendix:parameterComparison}e,f). 
Fourthly, a high-level control interval of $K=10$ steps (left-hand column of plots) increases the modulatory effect of the high-level noise compared to a control interval of $K=1$ step. For instance, for a K=1 step high-level control interval several of the snake controllers show little response to the high-level modulation independently of its strength; they largely maintain the stereotyped trajectory shape. This can be observed in Fig.\ \ref{fig:appendix:parameterComparison}b. In other cases stronger modulation appears to be required to compensate for the short control interval.

}
\end{document}